\documentclass[10pt,twocolumn,letterpaper]{article}

\usepackage{iccv}
\usepackage{times}
\usepackage{epsfig}
\usepackage{graphicx}
\usepackage{amsmath}
\usepackage{amssymb}
\usepackage{subfig}
\usepackage{tabularx}
\usepackage{booktabs}
\usepackage{multirow}
\usepackage{enumitem}
\usepackage{multicol}
\usepackage{csquotes}
\usepackage[dvipsnames, table]{xcolor}
\usepackage{color}
\usepackage{rotating}
\usepackage{floatrow}
\usepackage{array, makecell}
\usepackage[flushleft]{threeparttable} 
\usepackage{csquotes}
\usepackage{cite}
\usepackage{algorithm}
\usepackage{algpseudocode}
\usepackage{amstext} 
\usepackage{array}   
\newcolumntype{M}{>{$}c<{$}}

\newcommand{\bx}{\mathbf{x}}
\newcommand{\bh}{\mathbf{h}}
\newcommand{\bz}{\mathbf{z}}

\newcommand{\IR}{\mathbb{R}}
\newcommand{\IE}{\mathbb{E}}
\newcommand{\EL}{\mathcal{L}}
\newcommand{\ET}{\mathcal{T}}
\newcommand{\EF}{\mathcal{F}}
\newcommand{\EB}{\mathcal{B}}

\newcommand{\ED}{\mathcal{D}}

\newcommand{\EZ}{\mathcal{Z}}
\newcommand{\EP}{\mathcal{P}}

\DeclareRobustCommand{\eg} {\textit{e}.\textit{g}.}
\DeclareRobustCommand{\ie}{\textit{i}.\textit{e}.}
\DeclareRobustCommand{\etal}{\textit{et~al.}~}

\newcommand{\Section}[1]{\vspace{-1mm} \section{#1} \vspace{-1mm}}
\newcommand{\SubSection}[1]{\vspace{-1mm} \subsection{#1} \vspace{-1mm}}

\newcommand{\Paragraph}[1]{\vspace{1mm}\noindent\textbf{#1.}\hspace{0.5mm}}
\newcommand{\ParagraphMini}[1]{\vspace{0.8mm}\emph{#1.}\hspace{0.5mm}}

\newcommand{\firstkey}[1]{\textcolor{red}{\mathbf{#1}}}
\newcommand{\secondkey}[1]{\textcolor{blue}{\mathbf{#1}}}
\newcommand{\firstkeytext}[1]{\textcolor{red}{\textbf{#1}}}
\newcommand{\secondkeytext}[1]{\textcolor{blue}{\textbf{#1}}}

\usepackage[pagebackref=true,breaklinks=true,letterpaper=true,colorlinks,bookmarks=false]{hyperref}

\iccvfinalcopy 

\def\iccvPaperID{10033} 

\ificcvfinal\fi

\begin{document}

\title{Unified Detection of Digital and Physical Face Attacks}

\author{Debayan Deb, Xiaoming Liu, Anil K. Jain\\
Department of Computer Science and Engineering,\\
Michigan State University, East Lansing, MI, 48824\\
{\tt\small{\{debdebay, liuxm, jain\}@cse.msu.edu}}}

\ificcvfinal\thispagestyle{empty}\fi

\twocolumn[{%
\renewcommand\twocolumn[1][]{#1}%
\maketitle
\begin{center}
    \centering
    \captionsetup{font=small}
    \includegraphics[width=0.9\linewidth]{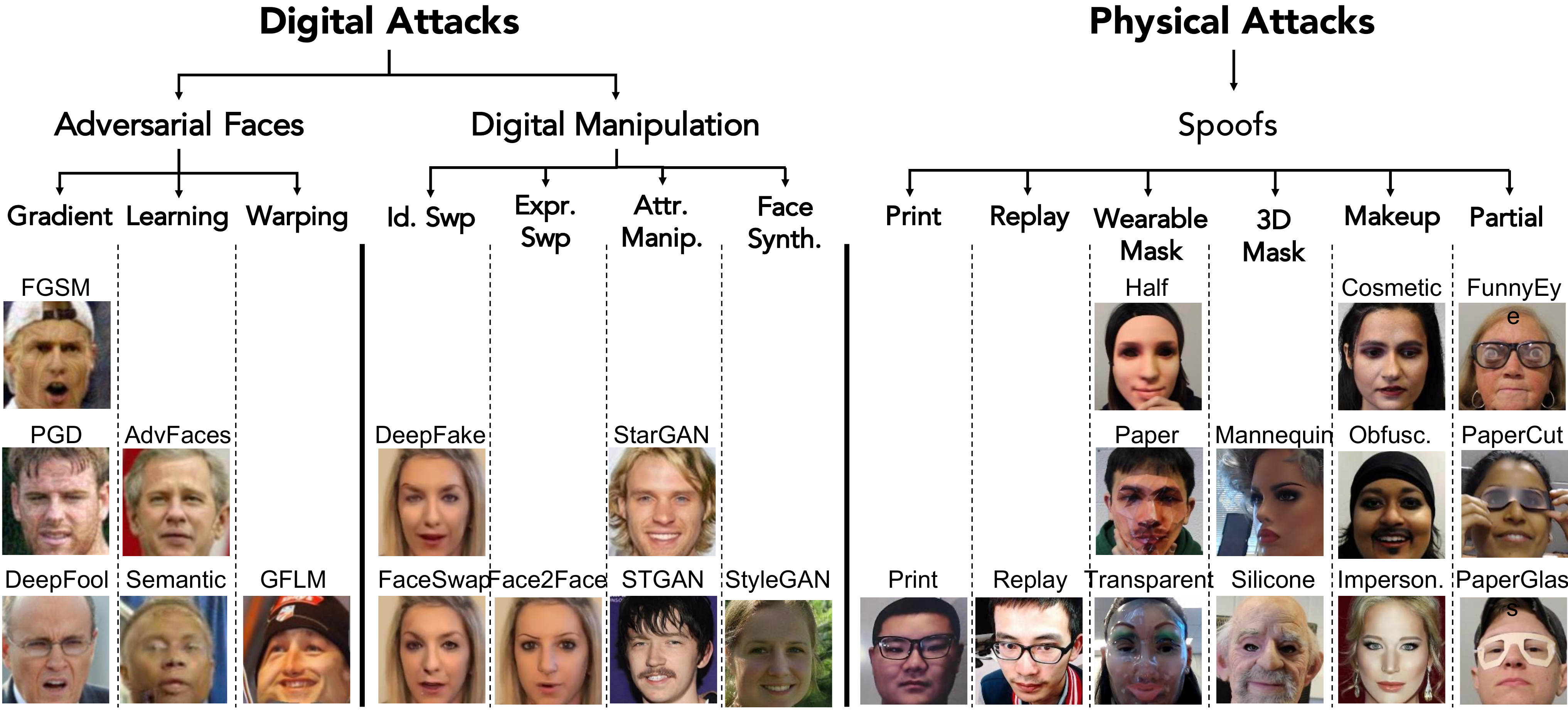}\vspace{-2mm}
    \captionof{figure}{Face attacks against AFR systems are continuously evolving in both digital and physical spaces. Given the diversity of the face attacks, prevailing methods fall short in detecting attacks across all three categories (\ie, adversarial, digital manipulation, and spoofs). This work is among the first to define the task of face attack detection on the $25$ attack types across $3$ categories shown here.}
    \label{fig:frontpage}
\end{center}%
}]

\begin{abstract}
   \vspace{-3mm}State-of-the-art defense mechanisms against face attacks achieve near perfect accuracies within one of three attack categories, namely adversarial, digital manipulation, or physical spoofs, however, they fail to generalize well when tested across all three categories. Poor generalization can be attributed to learning incoherent attacks jointly. To overcome this shortcoming, we propose a unified attack detection framework, namely UniFAD, that can automatically cluster 25 coherent attack types belonging to the three categories. Using a multi-task learning framework along with $k$-means clustering, UniFAD learns joint representations for coherent attacks, while uncorrelated attack types are learned separately. Proposed UniFAD outperforms prevailing defense methods and their fusion with an overall TDR = $94.73\%$ @ $0.2\%$ FDR on a large fake face dataset consisting of $341K$ bona fide images and $448K$ attack images of $25$ types across all 3 categories. Proposed method can detect an attack within $3$~milliseconds on a Nvidia 2080Ti. UniFAD can also identify the attack categories with $97.37\%$ accuracy. Code and dataset will be publicly available.
\end{abstract}

\vspace{-2mm}
\Section{Introduction}
\label{sec:intro}
Automated face recognition (AFR) systems have been projected to grow to USD $3.35$B by 2024\footnote{\url{https://bwnews.pr/2OqY0nD}}. It is estimated that over a billion smartphones today unlock via face authentication\footnote{\url{https://bit.ly/30vYBHg}}. However, the foremost challenge facing AFR systems is their vulnerability to {\it face attacks}. For instance, an attacker can hide his identity by wearing a 3D mask~\cite{3d_mask}, or intruders can assume a victim's identity by digitally swapping their face with the victim's face image~\cite{detection_digital}. With unrestricted access to the rapid proliferation of face images on social media platforms, launching attacks against AFR systems has become even more accessible. Given the growing dissemination of ``fake  news"  and  ``deepfakes"~\cite{protect_deep_fakes}, the research  community and social media platforms alike are pushing towards {\it generalizable} defense against continuously evolving and sophisticated face attacks.


In literature, face attacks can be broadly classified into three attack categories: (i) Spoof attacks: artifacts in the \emph{physical} domain (\eg, 3D masks, eye glasses, replaying videos)~\cite{siw_m}, (ii) Adversarial attacks: imperceptible noises added to probes for evading AFR systems~\cite{delving_adv_faces}, and (iii) Digital manipulation attacks: entirely or partially modified photo-realistic faces using generative models~\cite{detection_digital}. 
Within each of these categories, there are different attack types. For example, each spoof medium, \eg, 3D mask and makeup, constitutes one attack type, and there are $13$ common types of spoof attacks~\cite{siw_m}. Likewise, in adversarial and digital manipulation attacks, each attack model, designed by unique objectives and losses, may be considered as one attack type.
Thus, the attack categories and types form  a $2$-layer tree structure encompassing the diverse attacks (see Fig.~\ref{fig:frontpage}). Such a tree will inevitably grow in the future.

In order to safeguard AFR systems against these attacks, numerous face attack detection approaches have been proposed~\cite{detection_digital, ssrfcn, one_class_spoof, disentangling, generalized_cues_spoof}. Despite impressive detection rates, prevailing research efforts focus on a few attack types within {\it one} of the three attack categories. Since the exact type of face attack
may not be known~\emph{a priori}, a generalizable detector that can defend an AFR system against any of the three attack categories is of utmost importance.

Due to the vast diversity in attack characteristics, from glossy 2D printed photographs to imperceptible perturbations in adversarial faces, we find that learning a single {\it unified} network is inadequate. Even when prevailing state-of-the-art (SOTA) detectors are trained on all $25$ attack types, they fail to generalize well during testing. Via ensemble training, we comprehensively evaluate the detection performance on fusing decisions from three SOTA detectors that individually excel at their respective attack categories. However, due to the diversity in attack characteristics, decisions made by each detector may not be complementary and result in poor detection performance across all $3$ categories. 

This research is among the first to focus on detecting~\emph{all $25$ attack types} known in literature ($6$ adversarial, $6$ digital manipulation, and $13$ spoof attacks). 
Our approach consists of (i) automatically clustering attacks with similar characteristics into distinct groups, and (ii) a multi-task learning framework to learn salient features to distinguish between bona fides and coherent attack types, while early sharing layers learn a joint representation to distinguish bona fides from any generic attack.

This work makes the following contributions:
\setlist{topsep=0mm}
\begin{itemize}[noitemsep]
    \item Among the first to define the task of face attack detection on $25$ attack types across $3$ attack categories: adversarial faces, digital face manipulation, and spoofs. 
    \item A novel~\textbf{uni}fied~\textbf{f}ace~\textbf{a}ttack~\textbf{d}etection framework, namely~\emph{UniFAD}, that automatically clusters similar attacks and employs a multi-task learning framework to detect digital and physical attacks.
    \item Proposed~\emph{UniFAD} achieves SOTA detection performance, TDR = $94.73\%$ @ $0.2\%$ FDR on a large fake face dataset, namely~\emph{GrandFake}. To the best of our knowledge,~\emph{GrandFake} is the largest face attack dataset studied in literature in terms of the number of diverse attack types.
    \item Proposed~\emph{UniFAD} allows for further identification of the attack categories, \ie, whether attacks are adversarial, digitally manipulated, or contains physical spoofing artifacts, with a classification accuracy of $97.37\%$.
\end{itemize}

\begin{figure*}
    \centering
    \captionsetup{font=small}
    \includegraphics[width=0.85\linewidth]{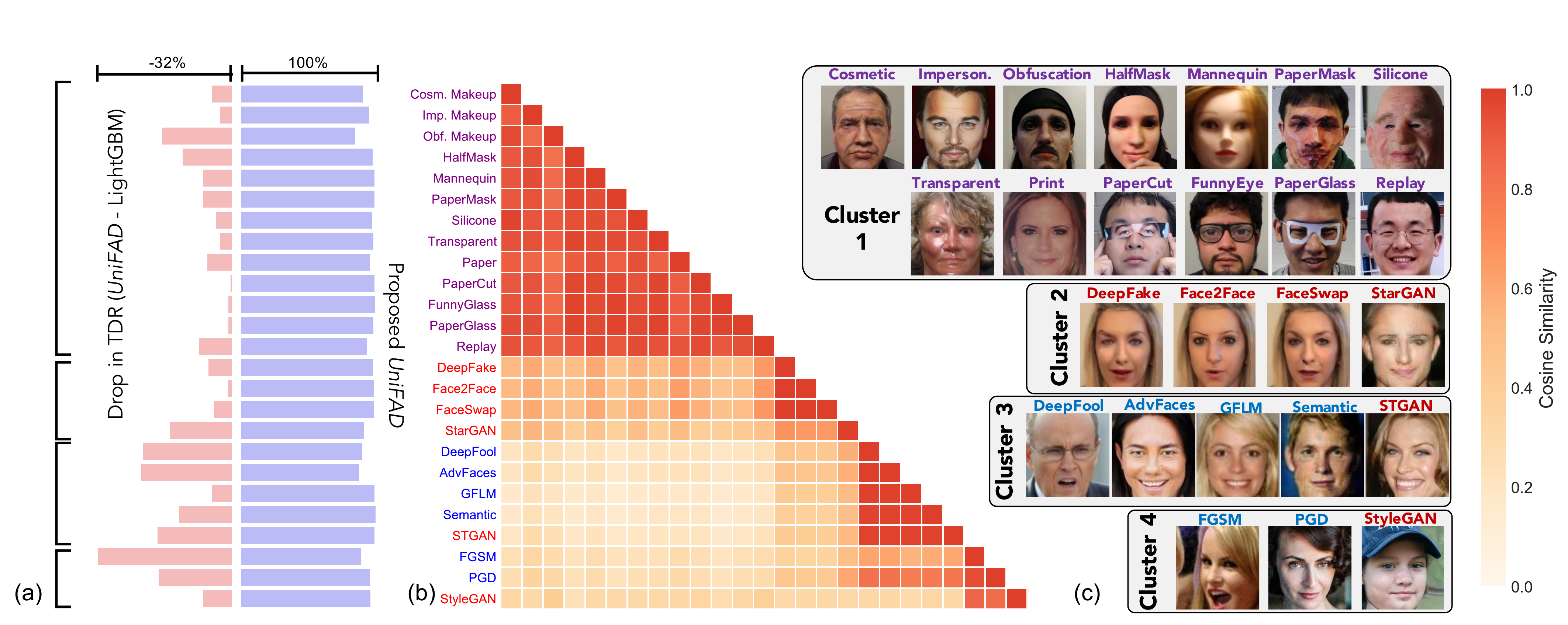}
    \vspace{-2mm}
    \caption{(a) Detection performance (TDR @ $0.2\%$ FDR) in detecting each attack type by the proposed~\emph{UniFAD} (purple) and the difference in TDR from the best fusion scheme, LightGBM~\cite{lightgbm} (pink). (b) Cosine similarity between mean features for $25$ attack types extracted by~\emph{JointCNN}. (c) Examples of attack types from $4$ different clusters via $k$-means clustering on JointCNN features. Attack types in purple, blue, and red denote spoofs, adversarial, and digital manipulation attacks, respectively.}
    \label{fig:attack_sim}
\end{figure*}

\begin{table}[!t]
\scriptsize
\setlength{\tabcolsep}{2.2pt}
\captionsetup{font=small}
\centering
\begin{threeparttable}
\renewcommand{\arraystretch}{1.2}
\begin{tabularx}{\linewidth}{l| l l M M M}
\noalign{\hrule height 1.0pt}
\multicolumn{1}{l}{} & \textbf{Study} & \textbf{Year} & \textbf{\# BonaFides} & \textbf{\# Attacks}  & \textbf{\# Types}\\
\noalign{\hrule height 1.0pt}

\parbox[t]{2mm}{\multirow{13}{*}{\raisebox{13.0em}{\rotatebox[origin=c]{90}{Adversarial}}}}
& UAP-D~\cite{uapd}  & 2018  & 9,959 & 29,877 & 1\\
& Goswami~\etal~\cite{goswami2019detecting} & 2019 & 16,685 & 50,055 & 3\\
& Agarwal~\etal~\cite{agarwal_image_transform} & 2020  & 24,042 & 72,126 & 3\\ 
& Massoli~\etal~\cite{massoli} & 2020 & 169,396 & 1M & 6\\ 
& FaceGuard~\cite{faceguard} & 2020 & 507,647 & 3M & 6\\ 
 \noalign{\hrule height 1.0pt}
\parbox[t]{2mm}{\multirow{14}{*}{\raisebox{13.0em}{\rotatebox[origin=c]{90}{Digital Manip.}}}}
& Zhou~\etal~\cite{zhou} & 2018  & 2,010 & 2,010 & 2\\
& Yang~\etal~\cite{yang_digital} & 2018 & 241 (I) / 49 (V) & 252 (I) / 49 (V) & 1\\
& DeepFake~\cite{deep_fake_face} & 2018  & - & 620 (V) & 1\\
& FaceForensics++~\cite{deep_fake_face} & 2019 & 1,000 (V) & 3,000 (V) & 3\\
& FakeSpotter~\cite{fake_spotter} & 2019   & 6,000 & 5,000 & 2\\
& DFFD~\cite{detection_digital} & 2020  & 58, 703 & 240,336  & 7\\
\noalign{\hrule height 1.0pt}
\parbox[t]{2mm}{\multirow{13}{*}{\raisebox{13.0em}{\rotatebox[origin=c]{90}{Phys. Spoofs}}}}
& Replay-Attack~\cite{replay_attack} & 2012  & 200 (V) & 1,000 (V) & 3\\
& MSU MFSD~\cite{msu_mfsd} & 2015  &   160 (V) & 280 (V) & 3\\
& OuluNPU~\cite{oulu_npu} & 2017  & 990 (V) & 3,960 (V) & 4\\
& SiW~\cite{auxiliary} & 2018  & 1,320 (V) & 3,158 (V) & 6\\
& SiW-M~\cite{siw_m} & 2019 & 660 (V) & 960 (V) & 13\\

\noalign{\hrule height 1pt}
\multicolumn{1}{l}{} & \textbf{\emph{GrandFake (ours)}} & \textbf{2021} & \mathbf{341,738} & \mathbf{447,674} & \mathbf{25}\\
\noalign{\hrule height 1pt}
\end{tabularx}
 \end{threeparttable}\vspace{-2mm}
\caption{Face attack datasets with no. of bona fide images, no. of attack images, and no. of attack types. Here, $I$ denotes images and $V$  refers to videos.~\emph{GrandFake} will be publicly available. } 
\label{tab:related_datasets}
\end{table}

\Section{Related Work}
\Paragraph{Individual Attack Detection} Early work on face attack detection primarily focused on one or two attack types in their respective categories. Studies on adversarial face detection\cite{gong,goswami2019detecting} primarily involved detecting gradient-based attacks, such as FGSM~\cite{fgsm}, PGD~\cite{projected_grad}, and DeepFool~\cite{deepfool}. DeepFakes were among the first studied digital attack manipulation~\cite{zhou, yang_digital, deep_fake_face}, however, generalizability of the proposed methods to a larger number of digital manipulation attack types is unsatisfactory~\cite{face_xray}. 
Majority of face anti-spoofing methods focus on print and replay attacks~\cite{active1, active3, active4, msu_mfsd, lbp1, yang, patch, deep_cnn, stasn, oulu_npu, auxiliary, silicone, active3,face-de-spoofing-anti-spoofing-via-noise-modeling}.

Over the years, a clear trend in the increase of attack types in each category can be observed in Tab.~\ref{tab:related_datasets}. Since a community of attackers dedicate their efforts to craft new attacks, it is imperative to comprehensively evaluate existing solutions against a large number of attack types.

\Paragraph{Joint Attack Detection} 
Recent studies have used multiple attack types in order to defend against face attacks. For \eg, FaceGuard~\cite{faceguard} proposed a generalizable defense against $6$ adversarial attack types. The Diverse Fake Face Dataset (DFFD)~\cite{detection_digital} includes $7$ digital manipulation attack types. In the spoof attack category, recent studies focus on detecting $13$ spoof types. 

Majority of the works tackling multiple attack types pose the detection as a binary classification problem with a single network learning a joint feature space. 
For simplicity, we refer to such a network architecture as~\emph{JointCNN}. 
For instance, it is common in adversarial face detection to train a JointCNN with bona fide faces and adversarial attacks synthesized on-the-fly by a generative network~\cite{faceguard, l2l, robgan, self_supervised}. 
On the other hand, majority of the proposed defenses against digital manipulation, fine-tune a pre-trained JointCNN (\eg, Xception~\cite{xception}) on bona fide faces and all available digital manipulation attacks~\cite{detection_digital, face_forensics, fake_spotter}. 
Due to the availability of a wide variety of physical spoof artifacts in face anti-spoofing datasets (\eg, eyeglasses, print and replay instruments, masks, \etc) along with evident cues for detecting them, studies on anti-spoofs are more sophisticated. The associated JointCNN employs either auxiliary cues, such as depth map and heart pulse signals (rPPG)~\cite{auxiliary, cdc, sun2020face}, or a ``compactness" loss to prevent overfitting~\cite{one_class_spoof, panoptic}.  
Recently Stehouwer {\it et al.}~\cite{noise-modeling-synthesis-and-classification-for-generic-object-anti-spoofing} attempt to learn a spoof detector from imagery of generic objects and apply it to face anti-spoofing. 
While jointly detecting multiple attack types is promising, detecting attack types~\emph{across} different categories is of the utmost importance. An early attempt proposed a defense against 4 attack types ($3$ spoofs and $1$ digital manipulation)~\cite{panoptic}. To the best of our knowledge, we are the first to attempt detecting $25$ attack types across $3$ categories.

\Paragraph{Multi-task Learning}
In multi-task learning (MTL), a task, $\ET_i$ is usually accompanied by a training dataset, $\ED_{tr}$ consisting of $N_t$ training samples, \ie, $D_{tr} = \{\bx_i^{tr}, y_i^{tr}\}_{i=1}^{N_{tr}}$, where $\bx_{i}^{tr} \in \IR$ is the $i$th training sample in $\ET_i$ and $y_i^{t}$ is its label. 
Most MTL methods rely on well-defined tasks~\cite{mtl1, mtl2,multi-task-convolutional-neural-network-for-pose-invariant-face-recognition}. Crawshaw~\etal~\cite{mtl_survey} summarize various works on MLT with CNNs. 
In this work, we propose a MTL framework in an extreme situation where only a single task is available (bona fide vs.~$25$ attack types) and utilize $k$-means clustering to construct new auxiliary tasks from $\ED_{tr}$. A recent study also utilized $k$-means for constructing new tasks, however, their approach utilizes a meta-learning framework where the groups themselves can alter throughout training~\cite{kmeans_mtl}. We show that this is problematic for face attacks since attacks that share similar characteristics should be trained jointly. Instead, we propose a new unified attack detection framework that first utilizes $k$-means to partition the $25$ attacks types, and then learns shared and attack-specific representations to distinguish them from bona fides.

\begin{figure*}
    \centering
    \captionsetup{font=small}
    \includegraphics[width=\linewidth]{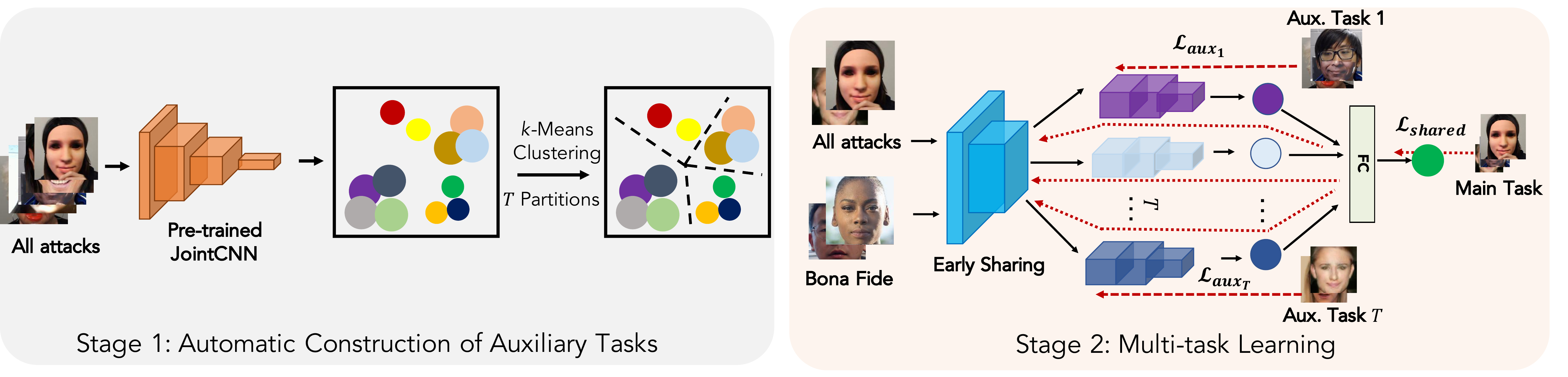}
    \vspace{-6mm}
    \caption{An overview of training~\emph{UniFAD} in two stages. Stage $1$ automatically clusters coherent attack types into $T$ groups. Stage $2$ consists of a MTL framework where early layers learn generic attack features while $T$ branches learn to distinguish bona fides from coherent attacks.} 
    \label{fig:overview}
\end{figure*}

\Section{Dissecting Prevailing Defense Systems}

\vspace{1mm}
\SubSection{Datasets} 
In order to detect $25$ attack types ($6$ adversarial, $6$ digital manipulation, and $13$ spoofs), we propose the~\emph{GrandFake} dataset, an amalgamation of multiple face attack datasets from each category. We provide additional details of~\emph{GrandFake} in Sec.~\ref{sec:datasets}.

\SubSection{Drawback of JointCNN}
Consider the diversity in the available attacks: from imperceptible adversarial perturbations to digital manipulation attacks, both of which are entirely different from physical print attacks (hard surface, glossy, 2D). Even within the spoof category, characteristics of mask attacks are quite different from replay attacks. In addition, discriminative cues for some attack types may be observed in high-frequency domain (\eg, defocused blurriness, chromatic moment), while others exhibit low-frequency cues (\eg, color diversity and specular reflection). For these reasons, learning a common feature space to discriminate all attack types from bona fides is challenging and a JointCNN may fail to generalize well even on attack types seen during training.

We demonstrate this by first training a JointCNN on the $25$ attack types in~\emph{GrandFake} dataset. We then compute an~\emph{attack similarity matrix} between the $25$ types (see Fig.~\ref{fig:attack_sim}(b)). The mean feature for each attack type is first computed on a validation set composed of $1,000$ images per attack. We then compute the pairwise cosine similarity between mean features from all attack pairs. 
From Fig.~\ref{fig:attack_sim}, we note that physical attacks have little correlation with adversarial attacks and therefore, learning them jointly within a common feature space may degrade detection performance. 

Although prevailing JointCNN-based defense achieve near perfect detection when trained and evaluated on the respective attack types in isolation, we observe a significantly degraded performance when trained and tested on all $3$ attack categories together (see Tab.~\ref{tab:detection}). In other words, even when a prevailing SOTA defense system is trained on all $3$ categories, it may lead to degraded performance on testing.

\SubSection{Unifying Multiple JointCNNs}
\label{sec:unifying}
Another possible approach is to consider ensemble techniques; instead of using a single JointCNN detector, we can fuse decisions from multiple individual detectors that are already experts in distinguishing between bona fides and attacks from their respective attack category. 
Given three SOTA detectors, one per attack category, we perform a comprehensive evaluation on parallel and sequential score-level fusion schemes. 


In our experiments, we find that, indeed, fusing score-level decisions from single-category detectors outperforms a single SOTA defense system trained on all attack types.  
Note that efforts in utilizing prevailing defense systems rely on the assumption that attack categories are independent of each other. However, Fig.~\ref{fig:attack_sim} shows that some digital manipulation attacks, such as STGAN and StyleGAN, are {\it more closely related} to some of the adversarial attacks (\eg, AdvFaces, GFLM, and Semantic) than other digital manipulation types. This is likely because all five methods utilize a GAN to synthesize their attacks and may share similar attack characteristics. 
Therefore, a SOTA adversarial detector and a SOTA digital manipulation detector may individually excel at their respective categories, but may not provide complementary decisions when fused. Instead of training detectors on groups with manually assigned semantics (\eg, adversarial, digital manipulation, spoofs), it is better to train JointCNNs on coherent attacks.  In addition, utilizing decisions from pre-trained JointCNNs may tend to overfit to the attack categories used for training. 

\Section{Proposed Method: UniFAD}
We propose a new multi-task learning framework for Unified Attack Detection, namely~\emph{UniFAD}, by training an end-to-end network for improved physical and digital face attack detection. In particular, a $k$-means augmentation module is utilized to automatically construct auxiliary tasks to enhance single task learning (such as a JointCNN). 
Then, a joint model is decomposed into a feature extractor (shared layers) $\EF$ that is shared across all tasks, and task-specific branches for each auxiliary task. Fig.~\ref{fig:overview} illustrates the auxiliary task creation and the training process of~\emph{UniFAD}.



\SubSection{Problem Definition} 
Let the ``main task" be defined as the overall objective of a unified attack detector: given an input image, $\bx$, assign a score close to $0$ if $\bx$ is bona fide or close to $1$ if $\bx$ is any of the available face attack types. 
We are also given a labeled training set, $D_{tr}$. Prevailing defenses follow a single task learning approach where the main task is adopted to be the ultimate training objective. In order to avoid the shortcomings of a JointCNN and unification of multiple JointCNNs, we first use $D_{tr}$ to automatically construct multiple auxiliary tasks $\{\ET_t\}_{t=1}^{T}$, where $T_i$ is the $i$th cluster of coherent attack types. If the auxiliary tasks are appropriately constructed, jointly learning these tasks along with the main task should improve unified attack detection compared to a single task learning approach.

\SubSection{Automatic Construction of Auxiliary Tasks}



One way to construct auxiliary tasks is to train a separate binary JointCNN on each attack type. 
Such partitioning massively increases computational burden (\eg, training and testing $25$ JointCNNs). 
Other simple partitioning methods, such as randomly partition are likely to cluster uncorrelated attacks. On the other hand, clustering in the pixel-space is also unappealing due to poor correlation between the distances in the pixel-space, and clustering in the high-dimensional space is challenging~\cite{mtl_pixel_space}. 
Therefore, we require a reasonable alternative to manual inspection of the attack similarity matrix in Fig.~\ref{fig:attack_sim} to partition the attack types into appropriate clusters. 

Fortunately, we already have a JointCNN trained via a single task learning framework that can extract salient representations. 
Thus, we can map the data $\{\bx\}$ into JointCNN's embedding space $\EZ$, producing $\{\bz\}$. We can then utilize a traditional clustering algorithm, $k$-means, which takes a set of feature vectors as input, and clusters them into $k$ distinct groups based on a geometric constraint. Specifically, for each attack type, we first compute the mean feature. We then utilize $k$-means clustering to partition the $L$ features into $T (\leq L)$ sets, $\EP = \{\EP_1, \EP_2, \ldots, \EP_{T}\}$ such that within-cluster sum of squares (WCSS) is minimized, 
\begin{align}
    \arg\min_{\EP}\sum_{i=1}^{T}\sum_{\bar\bz \in \EP_{i}} || \bar\bz - \mu_{i} ||^2, 
\end{align}
where, $\bar\bz$ represents a mean feature for an attack type and $\mu_{i}$ is the mean of the features in $\EP_i$. Fig.~\ref{fig:attack_sim}(c) shows an example on clustering the $25$ attack types of~\emph{GrandFake}. 

\SubSection{Multi-Task Learning with Constructed Tasks}
With a multi-task learning framework, we learn coherent attack types jointly, while uncorrelated attacks are learned in their own feature spaces. We construct $T$ ``branches" where each branch is a neural network trained on a binary classification problem (\ie, aux. task). The learning objective of each branch, $\EB_{t}$, is to minimize,
\begin{align}
    \EL_{aux_{t}} = \IE_{\bx}\left[\text{log}\EB_{t}(\bx_{bf})\right] + \IE_{\bx}\left[\text{log}\left(1 - \EB_{t}(\bx_{fake}^{\EP_{t}})\right)\right].
\end{align}
where $\bx_{bf}$ denotes bona fide images and $\bx_{fake}^{\EP_{t}}$ is face attacks corresponding to the attack types in the partition $\EP_{t}$. 



\SubSection{Parameter Sharing}
\ParagraphMini{Early Sharing} We adopt a hard parameter sharing module which learns a common feature representation for distinguishing between bona fides and attacks prior to aux. task learning branches. Baxter~\cite{baxter} demonstrated the shared parameters have a lower risk of overfitting than the task-specific parameters.
Therefore, adopting early convolutional layers as a pre-processing step prior to branching can help~\emph{UniFAD} in its generalization to all $3$ categories. We construct hidden layers between the input and the branches to obtain shared features, $\bh = \EF(\bx)$, while the auxiliary learning branches output $\EB_{t}(\bh)$. 

\ParagraphMini{Late Sharing} Each branch $\EB_t$ is trained to output a decision score where scores closer to $0$ indicate that the input image is a bona fide, whereas, scores closer to $1$ correspond to attack types pertaining to the branch's partition. 
The scores from all $T$ branches are then concatenated and passed to a final decision layer. For simplicity, we define the final decision output as, $FC(\bx) := FC(\EB_{1}(\bh), \EB_{2}(\bh),\ldots, \EB_{T}(\bh))$.

The early shared layers and the final decision layer are learned via a binary cross-entropy loss,
\begin{align}
    \EL_{shared} = \IE_{\bx}\left[\text{log}FC(\bx_{bf})\right] + \IE_{\bx}\left[\text{log}\left(1 - FC(\bx_{fake})\right)\right],
\end{align}
between bona fides and all available attack types.

\SubSection{Training and Testing}
\label{sec:train_test}
The entire network is trained in an end-to-end manner by minimizing the following composite loss,
\begin{align}
    \EL_{UniFAD} = \EL_{shared} + \sum_{t=1}^{T}\EL_{aux_t}.
\end{align}
The $\EL_{shared}$ loss is backpropagated throughout~\emph{UniFAD}, while $\EL_{aux_t}$ is only responsible for updating the weights of the branch, $\EB_{t}$, and the final classification layer. For the forward and backward passes of $\EL_{shared}$, an equal number of bona fide and attack samples are used for training. On the other hand, for training each branch, $\EB_{t}$, we sample the equal number of bona fides and equal number of attack images from the attack partition $\EP_{t}$.

\begin{table*}[!t]
\centering
\footnotesize
\captionsetup{font=small}
\begin{threeparttable}
\begin{tabular}{l|lMcMMMMM}
\toprule
\multicolumn{2}{l}{\textbf{TDR (\%) @ 0.2\% FDR}} & \textbf{Year} & \textbf{Proposed For} & \textbf{Adv.} & \textbf{Dig. Man.} & \textbf{Phys.} & \textbf{Overall} & \textbf{Time (ms)}\\
\midrule
\parbox[t]{0.3mm}{\multirow{12}{*}{\raisebox{13.0em}{\rotatebox[origin=c]{90}{w/o Re-train}}}}
& FaceGuard~\cite{faceguard} & 2020 & Adversarial & \firstkey{99.91} & 22.28 & 00.58 & 29.64 & 01.41\\
& FFD~\cite{detection_digital} & 2020 & Digital Manipulation & 09.49 & \secondkey{94.57} & 01.25 & 34.55 & 11.57\\
& SSRFCN~\cite{ssrfcn} & 2020 & Spoofs & 00.25 & 00.76 & 93.19 & 22.71 & 02.22\\
& MixNet~\cite{mixnet} & 2020 & Spoofs & 00.36 & 09.83 & 78.21 & 21.12 & 12.47\\
\midrule
\parbox[t]{0.3mm}{\multirow{15}{*}{\raisebox{13.0em}{\rotatebox[origin=c]{90}{Baselines}}}}
& FaceGuard~\cite{faceguard} & 2020 & Adversarial  &  \secondkey{99.86} & 41.56 & 04.35 & 56.69 & 01.41\\
& FFD~\cite{detection_digital}  & 2020 & Digital Manipulation  & 76.06 & 91.32 & 87.43 & 68.25 & 11.57\\
& SSRFCN~\cite{ssrfcn} & 2020 & Spoofs  & 08.23 & 27.67 & 89.19 & 43.26 & 02.22 \\
& One-class~\cite{one_class_spoof} & 2020 & Spoofs  & 04.81 & 45.96 & 79.32 & 39.40 & 07.92\\
& MixNet-\emph{UniFAD} & 2021 & All & 82.33 & 91.59 & \secondkey{94.60} & \secondkey{90.07} & 12.47\\
\midrule
\parbox[t]{0.3mm}{\multirow{16}{*}{\raisebox{13.0em}{\rotatebox[origin=c]{90}{Fusion Schemes}}}}
& Cascade~\cite{viola_jones} & - & $-$ &  88.39 & 81.98 & 69.19 & 77.46 & 05.16\\
& Min-score & - & $-$ & 03.65 & 11.08 & 00.43 & 07.22 & 16.14\\
& Median-score & - & $-$ & 10.87 & 42.33 & 47.19 & 39.48 & 16.12\\
& Mean-score & - & $-$ & 14.53 & 47.18 & 61.32 & 38.23 & 16.12\\
& Max-score & - & $-$ & 85.32 & 61.93 & 56.87 & 73.89 & 16.13\\
& Sum-score & - & $-$ & 74.93 & 58.01 & 50.34 & 69.21 & 16.11\\
& LightGBM~\cite{lightgbm} & - & $-$ & 76.25 & 81.28 & 88.52  & 85.97 & 17.92\\
\midrule
\multicolumn{1}{l}{} & \emph{Proposed UniFAD} & 2021 & All & 92.56 & \firstkey{97.21} & \firstkey{98.76} & \firstkey{94.73} & 02.59\\
\bottomrule
\end{tabular}
\end{threeparttable}\vspace{-2mm}
\caption{Detection accuracy (TDR (\%) @ $0.2\%$ FDR) on~\emph{GrandFake} dataset. Results on fusing  FaceGuard~\cite{faceguard}, FFD~\cite{detection_digital}, and SSRFCN~\cite{ssrfcn} are also reported. We report time taken to detect a single image (on a Nvidia 2080Ti GPU). [Keys: \firstkeytext{Best}, \secondkeytext{Second best}]}
\label{tab:detection}
\end{table*}


\ParagraphMini{Attack Detection} During testing, an image passes through the shared layers and then each branch of~\emph{UniFAD} outputs a decision whether the image is bona fide (values close to $0$) or an attack (close to $1$). The final decision layer outputs the final decision score. Unless stated otherwise, we use the final decision scores to report  performance.

\ParagraphMini{Attack Classification} Once an attack is detected,~\emph{UniFAD} can automatically classify the attack type and category. For all $L$ attack types in the training set, we extract intermediate $128$-dim feature vectors from $T$ branches. The features are then concatenated and the mean feature across all $L$ attack types is computed, such that, we have $L$ feature vectors of size $T\times128$. For a detected attack, Cosine similarity is computed between the testing sample's feature vector and the mean training features for $L$ types. The predicted attack type is the one with the highest similarity score.


\Section{Experimental Results} 
\vspace{1mm}
\SubSection{Experimental Settings}
\Paragraph{Dataset}\label{sec:datasets} 
\emph{GrandFake} consists of $25$ face attacks from $3$ attack categories. 
Both bona fide and fake faces are of varying quality due to different capture conditions.

\ParagraphMini{Bona Fide Faces} We utilize faces from CASIA-WebFace~\cite{casia}, LFW~\cite{lfw}, CelebA~\cite{celeba}, SiW-M~\cite{siw_m}, and FFHQ~\cite{ffhq} datasets since the faces therein cover a broad variation in race, age, gender, pose, illumination, expression, resolution, and acquisition conditions.

\ParagraphMini{Adversarial Faces} We craft adversarial faces from CASIA-WebFace~\cite{casia} and LFW~\cite{lfw} via $6$ SOTA adversarial attacks: FGSM~\cite{fgsm}, PGD~\cite{projected_grad}, DeepFool~\cite{deepfool}, AdvFaces~\cite{advfaces}, GFLM~\cite{gflm}, and SemanticAdv~\cite{semantic_adv}. 
These attacks were chosen for their success in evading SOTA AFR systems such as ArcFace~\cite{arcface}. 

\ParagraphMini{Digital Manipulation} 
There are four broad types of digital face manipulation: identity swap, expression swap, attribute manipulation, and entirely synthesized faces~\cite{detection_digital}. 
We use all clips from FaceForensics++~\cite{face_forensics},  
including identity swap by FaceSwap and DeepFake\footnote{\url{https://github.com/deepfakes/faceswap}}, and expression swap by Face2Face~\cite{face2face}.
We utilize two SOTA models, StarGAN~\cite{stargan} and STGAN~\cite{stgan}, to generate attribute manipulated faces in Celeba~\cite{celeba} and FFHQ~\cite{ffhq}. 
We use the pretrained StyleGAN2 model\footnote{\url{https://github.com/NVlabs/stylegan2}} to synthesize $100$K fake faces. 

\ParagraphMini{Physical Spoofs} We utilize the publicly available \emph{SiW-M} dataset~\cite{siw_m}, comprising $13$ spoof types. Compared with other spoof datasets (Tab.~\ref{tab:related_datasets}), SiW-M is diverse in spoof attack types, environmental conditions, and face poses.

\Paragraph{Protocol} As is common practice in face recognition literature, bona fides and attacks from CASIA-WebFace~\cite{casia} are used for training, while bona fides and attacks for LFW~\cite{lfw} are sequestered for testing. The bona fides and attacks from other datasets are split into $70\%$ training, $5\%$ validation, and $25\%$ testing. 


\Paragraph{Implementation}~\emph{UniFAD} is implemented in Tensorflow, 
and trained with a constant learning rate of $(1e^{-3})$ with a mini-batch size of $180$. The objective function, $\EL_{UniFAD}$, is minimized using Adam optimizer for $100K$ iterations. 
Data augmentation during training involves random
horizontal flips with a probability of $0.5$.

\Paragraph{Metrics} Studies on different attack categories provide their own metrics. Following the recommendation from IARPA ODIN program, we report the TDR @ $0.2\%$ FDR\footnote{\url{https://www.iarpa.gov/index.php/research-programs/odin}} and the overall detection accuracy (in Supp.).


\SubSection{Comparison with Individual SOTA Detectors}

In this section, we compare the proposed~\emph{UniFAD} to prevailing face attack detectors via publicly available repositories provided by the authors (see Supp.).

\ParagraphMini{Without Re-training} In Tab.~\ref{tab:detection}, we first report the performance of $4$ pre-trained SOTA detectors. These baselines were chosen since they report the best detection performance in datasets with largest numbers of attack types in their respective categories (see Tab.~\ref{tab:related_datasets}). We also report the performance of a generalizable spoof detector with sub-networks, namely~\emph{MixNet}. MixNet semantically assigns spoofs into groups (print, replay, and masks) for training each sub-network without any shared representation.
We find that pre-trained methods indeed excel in their specific attack categories, however, generalization performance across all $3$ categories deteriorates catastrophically.

\ParagraphMini{With Re-training} After re-training the $4$ SOTA detectors on all $25$ attack types, we find that they generalize better across categories. FaceGuard~\cite{faceguard}, FFD~\cite{detection_digital}, SSRFCN~\cite{ssrfcn}, and One-Class~\cite{one_class_spoof}, all employ a JointCNN for detecting attacks. Unsurprisingly, these defenses perform well on some attack categories, while failing on others.

For a fair comparison, we also modify MixNet, namely~\emph{MixNet-UniFAD} such that clusters are assigned via $k$-means with $4$ branches. In contrast to~\emph{MixNet-UniFAD},~\emph{UniFAD} (i) employs early shared layers for generic attack cues, and (ii) each branch learns to distinguish between bona fides and specific attack types. MixNet, on the other hand, assigns a bona fide label ($0$) to attack types outside a respective branch's partition. This negatively impacts network convergence. Overall, we find that~\emph{UniFAD} outperforms~\emph{MixNet-UniFAD} with TDR $90.07\% \xrightarrow{} 94.73\%$ @ $0.2\%$ FDR.

\begin{figure}
    \captionsetup{font=small}
    \centering
    \includegraphics[width=0.85\linewidth]{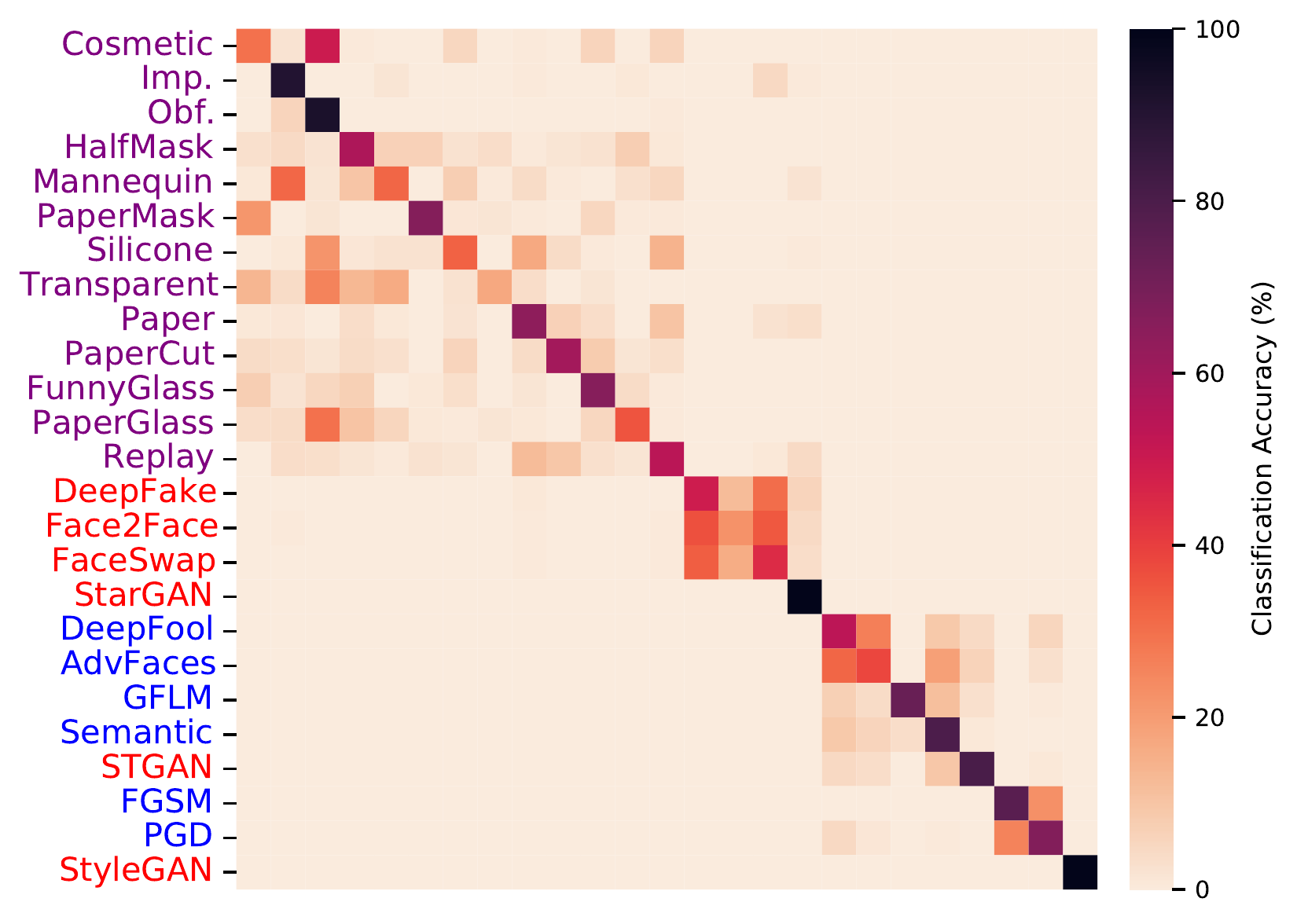}\label{fig:cm_attack}\vspace{-2mm}
    \caption{Confusion matrix representing the classification accuract of~\emph{UniFAD} in identifying the $25$ attack types. Majority of misclassifications occur within the attack category. Darker values indicate higher accuracy. Overall,~\emph{UniFAD} achieves $75.81\%$ and $97.37\%$ classification accuracy in identifying attack types and categories, respectively. Purple, blue, and red denote spoofs, adversarial, and digital manipulation attacks, respectively.}
    \label{fig:classification}
\end{figure}

\SubSection{Comparison with Fused SOTA Detectors}
We also comprehensively evaluate detection performance on fusing SOTA detectors. We utilize three best performing detectors from each attack category, namely FaceGuard~\cite{faceguard}, FFD~\cite{detection_digital}, and SSRFCN~\cite{ssrfcn}. Inspired by the Viola-Jones object detection~\cite{viola_jones}, we adopt a sequential ensemble technique, namely Cascade~\cite{viola_jones}, where an input probe is passed through each detector sequentially. We also evaluate $5$ parallel score fusion rules (min, mean, median, max, and sum) and a SOTA ensemble technique, namely LightGBM~\cite{lightgbm}. More details are provided in Supp. Indeed, we observe an overhead in detection speed compared to the individual detectors in isolation, however, cascade, max-score fusion and LightGBM~\cite{lightgbm} can enhance the overall detection performance compared to the individual detectors at the cost of slower inference speed. Since the individual detectors still train with incoherent attack types, we find that proposed~\emph{UniFAD} outperforms all the considered fusion schemes.

In Fig.~\ref{fig:attack_sim}(a), we show the performance degradation of LightGBM~\cite{lightgbm}, the best fusing baseline, w.r.t.~\emph{UniFAD}.
We observe that among $4$ clusters, the last $2$ have the overall largest degradation.
Interestingly, these $2$ clusters are the only ones including attack types across different attack categories, learned via our $k$-mean clustering.
In other words, the cross-category attacks types within a branch benefit each other, leading to the largest performance gain over~\cite{lightgbm}. 
This further demonstrates the necessity and importance of a unified detection scheme --- the more attack types the detector sees, the more likely it would nourish among each other and be able to generalize.

\SubSection{Attack Classification} 
We classify the exact attack type and categories using the method described in Sec.~\ref{sec:train_test}.
In Fig.~\ref{fig:classification}, we find that~\emph{UniFAD} can predict the attack type with $75.81\%$ classification accuracy. While predicting the exact type may be challenging, we highlight that majority of the misclassifications occurs within attack's category. 
Without human intervention, once~\emph{UniFAD} is deployed in AFR pipelines, it can predict whether an input image is adversarial, digitally manipulated, or contains spoof artifacts with $97.37\%$ accuracy.

\SubSection{Analysis of UniFAD}

\begin{figure}
\vspace{-3mm}
\captionsetup{font=small}
  \captionsetup[subfigure]{labelformat=empty}
    \centering
    \subfloat[]{\includegraphics[width=0.5\linewidth]{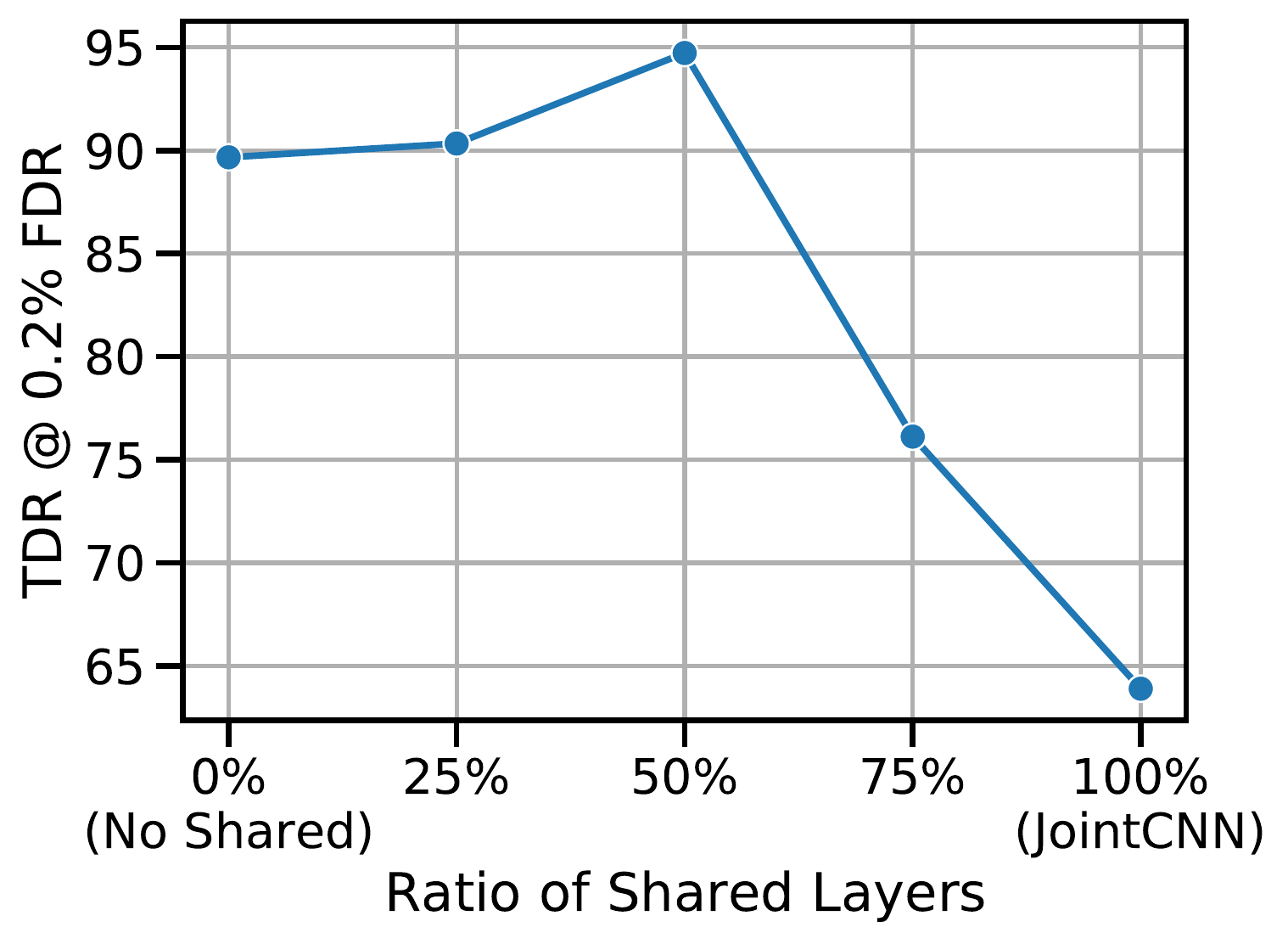}\label{fig:perf_v_shared}}
    \subfloat[]{\includegraphics[width=0.5\linewidth]{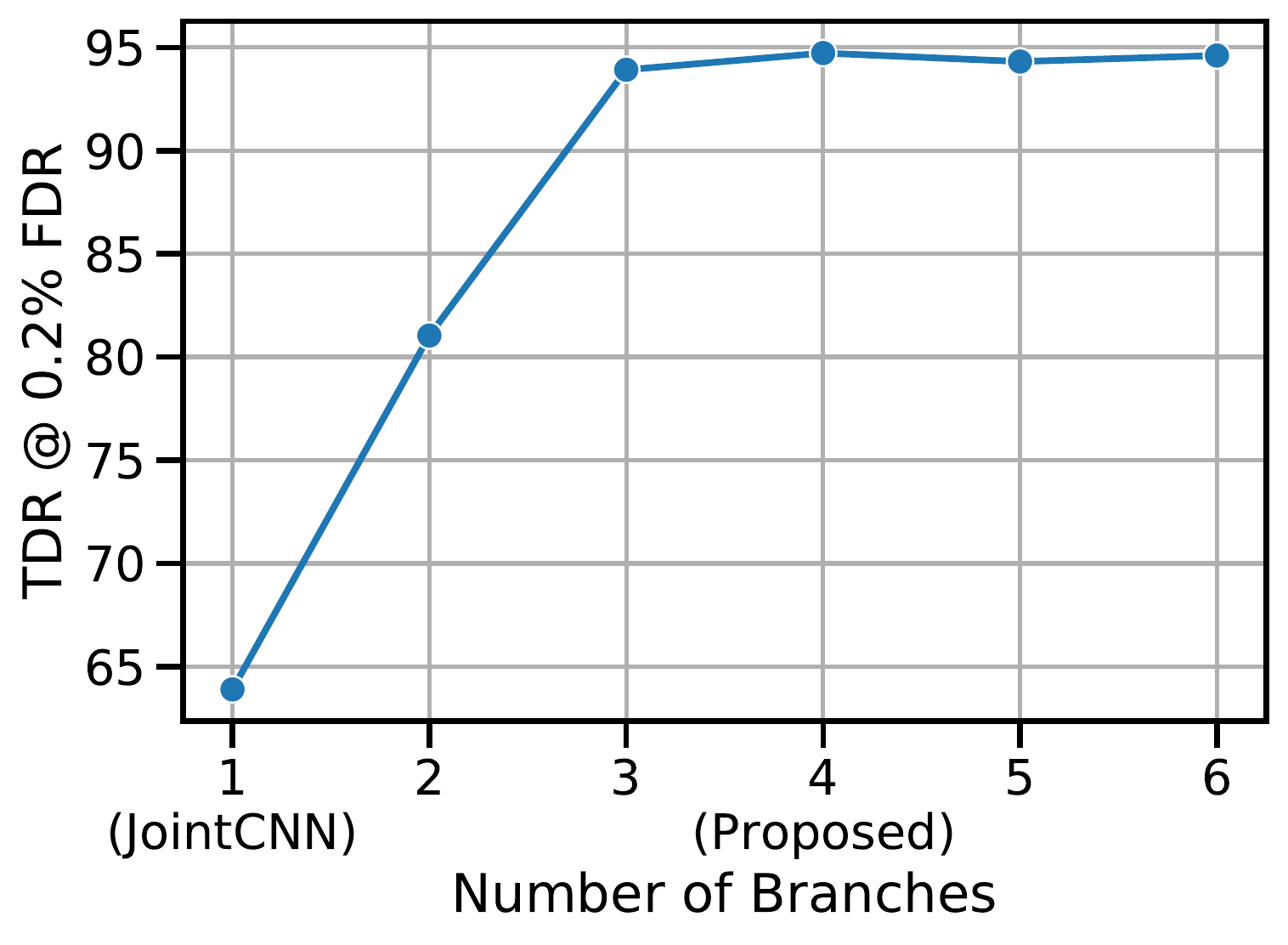}\label{fig:perf_v_clusters}}\vspace{-6mm}
    \caption{Detection performance with respect to varying ratio of shared layers (left) and number of branches (right). Our proposed architecture uses $50\%$ shared layers with $4$ branches.}
    \label{fig:analysis_mtl}
\end{figure}
\vspace{-1mm}
\Paragraph{Architecture} We first ablate and analyze our architecture.

\ParagraphMini{Ratio of Shared Layers} Our backbone network consists of a $4$-layer CNN. In Fig.~\ref{fig:perf_v_shared}, we report the detection performance when we incorporate $0$, $1$ ($25\%$), $2$ ($50\%$), $3$ ($75\%$), and $4$ ($100\%$) layers for early sharing. We observe a trade-off between detection performance and the number of early layers: too many reduces the effects of learning task-specific features via branching, whereas, less number of shared layers inhibits the network from learning generic features that distinguish any attack from bona fides. We find that an even split results in superior detection performance.

\ParagraphMini{Number of Branches} In Fig.~\ref{fig:perf_v_clusters}, we vary the number of branches (aux.~tasks constructed via $k$Means) and report the detection performance. Indeed, increasing the number of branches via additional clusters enhances detection performance. However, the performance saturates after $4$ branches.~\emph{UniFAD} with $4$ branches achieves TDR = $94.73\%$ @ $0.2\%$ FDR, whereas, $5$ and $6$ branches achieve TDRs of $94.33$ and $94.62$ at $0.2\%$, respectively. For $T=5$, we notice that StyleGAN is isolated from Cluster $3$ (see Fig.~\ref{fig:attack_sim}(c)) into a separate cluster. Learning to discriminate StyleGAN separately may offer no significant advantage than learning jointly with FGSM and PGD. Thus, we choose $T=4$ due to lower network complexity and higher inference efficiency. 

\begin{figure}
    \captionsetup{font=small}
    \centering
    \includegraphics[width=0.95\linewidth]{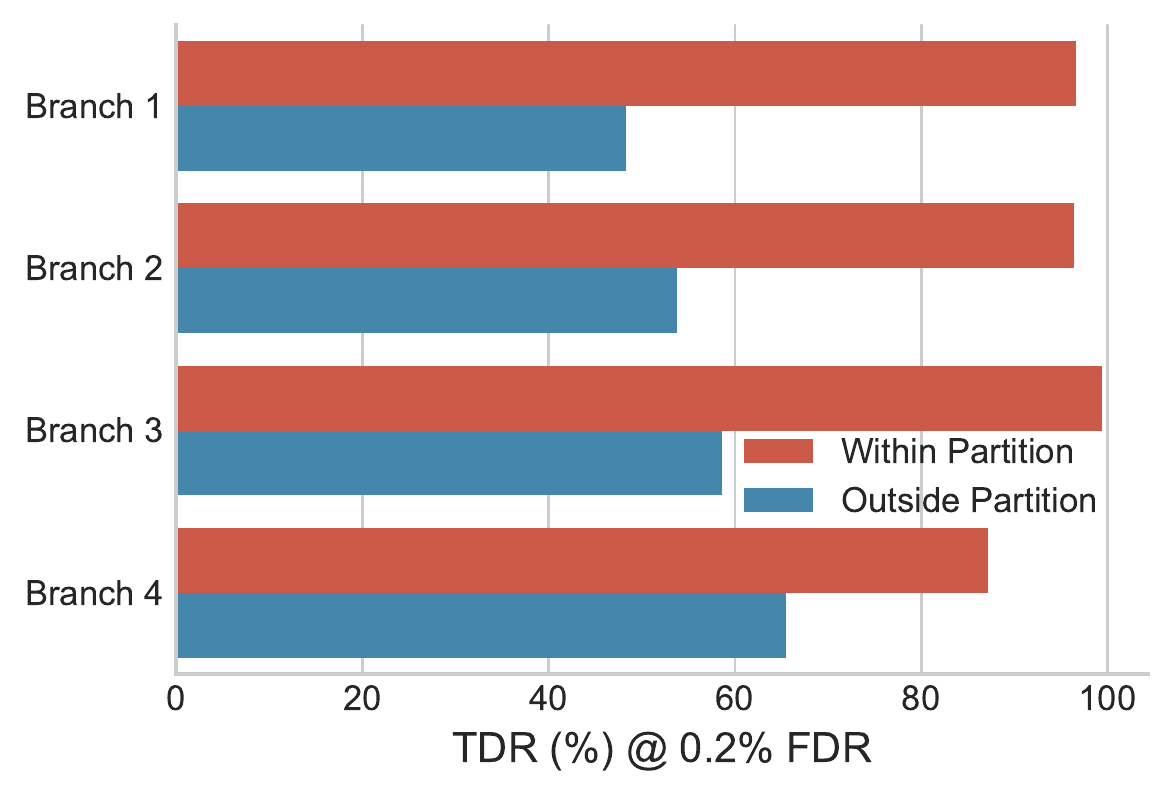}\vspace{-2mm}
    \caption{Detection performance on attack types within and outside a branch's partition. Performance drops on attacks outside partition as they may not have any correlation with within-partition attack types.}\vspace{-2mm}
    \label{fig:branch_perf}
\end{figure}

\ParagraphMini{Branch Generalizability} In Fig.~\ref{fig:branch_perf}, scores from the $4$ branches are used to compute the detection performance on attack types within respective partitions and those outside a branch's partition (see Fig.~\ref{fig:attack_sim}(b)). 
Since attack types outside a branch's partition are purportedly incoherent, we see a drop in performance; validating the drawback of JointCNN. We find that the lowest performance branch, Branch $4$, also exhibits the best generalization performance across other attack types. This is likely because learning to distinguish bona fides from imperceptible perturbations from FGSM, PGD, and minute synthetic noises from StyleGAN yields a tighter decision boundary which may contribute to better generalization across digital attacks. Anti-spoofing (Branch $1$) itself does not directly aid in detecting digital attacks.

\begin{table}[t!]
\captionsetup{font=small}
\setlength{\tabcolsep}{2.0pt}
\begin{center}
\scriptsize
\begin{tabularx}{1.00\linewidth}{X || c|c|c || M}
\Xhline{2\arrayrulewidth}
\multirow{2}{*}{Model} & \multicolumn{3}{c||}{Modules}  & \text{Overall}\\
\cline{2-5}
& Shared Layers & Branching & $k$Means & \text{TDR (\%) @ $0.2\%$ FDR} \\\Xhline{2\arrayrulewidth}
JointCNN & \checkmark & & & 63.89 \\ \hline
$\EB_{Semantic}$ &  & \checkmark &  & 86.17  \\ \hline
$\EB_{Random}$ &  & \checkmark  &  & 53.95 \pm 08.02 \\ \hline
$\EB_{kMeans}$ &  & \checkmark & \checkmark &  89.67 \\ \hline
SharedSemantic & \checkmark & \checkmark & & 92.44 \\ \hline
\emph{Proposed} & \checkmark  & \checkmark & \checkmark & \mathbf{94.73} \\
\Xhline{2\arrayrulewidth}
\end{tabularx}
\vspace{-5mm}
\caption{Ablation study over components of~\emph{UniFAD}. Branching via ``$\EB_{Semantic}$", ``$\EB_{Random}$", and ``$\EB_{kMeans}$" refer to partitioning attack types by their semantic categories, randomly, and $k$Means. ``SharedSemantic" includes shared layers prior to branching.}
\label{tab:ablation}
\end{center}
\end{table}

\Paragraph{Ablation Study} In Tab.~\ref{tab:ablation}, we conduct a component-wise ablation study over~\emph{UniFAD}. We study different partitioning techniques to group the $25$ attack types. We employ semantic partitioning, $\EB_{Semantic}$ where attack types are clustered into the $3$ categories. Another technique is to split the $25$ attack types into $4$ clusters randomly, $\EB_{Random}$. We report the mean and standard deviation across $3$ trials of random splitting. We also report the performance of clustering via $k$Means. We find that both $\EB_{Semantic}$ and $\EB_{kMeans}$ outperforms JointCNN. Thus, learning separate feature spaces via MTL for disjoint attack types can improve overall detection compared to a JointCNN. We als find that incorporating early shared layers into $\EB_{Semantic}$, namely  $\EB_{SharedSemantic}$, can further improve detection from $86.17\% \xrightarrow{} 92.44\%$ TDR @ $0.2\%$ FDR. However, as we observed in Fig.~\ref{fig:attack_sim}, even within semantic categories, some attack types may be incoherent. By automatic construction of auxiliary tasks with $k$-means clustering and shared representation (Proposed), we can further enhance the detection performance to TDR = $94.73\%$ @ $0.2\%$ FDR.

\begin{figure}
    \centering
    \captionsetup{font=small}
    \includegraphics[width=\linewidth]{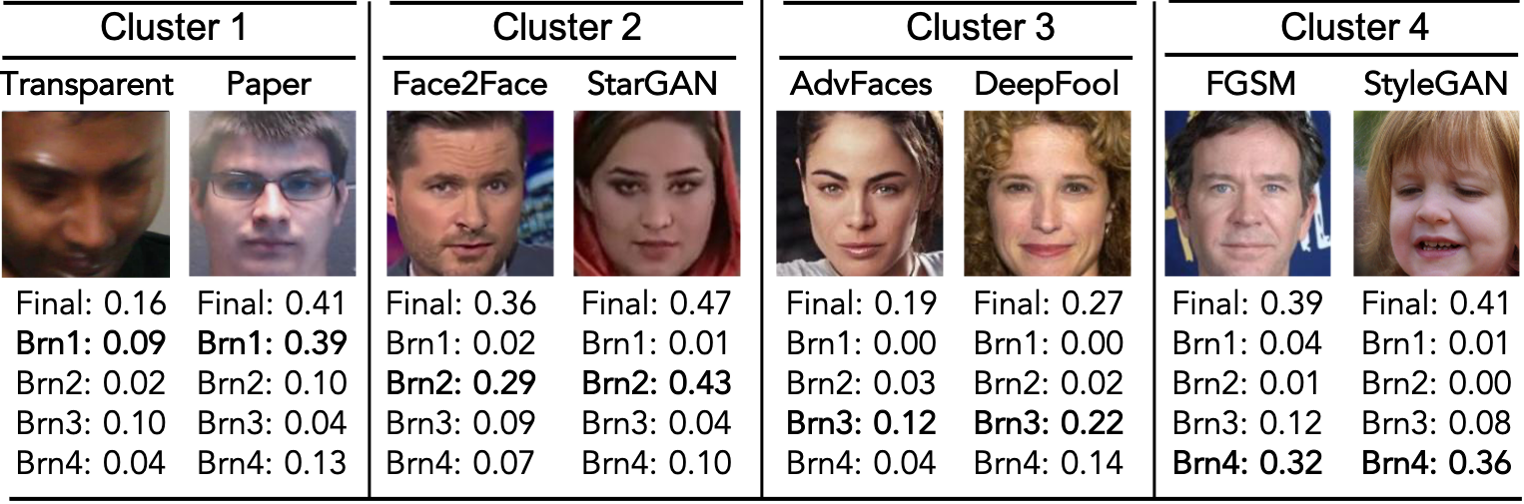}\vspace{-2mm}
    \caption{Example cases where~\emph{UniFAD} fails to detect face attacks. Final detection scores along with scores from each of the four branches ($\in [0,1]$) are given below each image. Scores closer $0$ indicate bona fide. Branches responsible for the respective cluster are highlighted in bold.}
    \label{fig:failure_cases}
\end{figure}

\Paragraph{Failure Cases} Fig.~\ref{fig:failure_cases} shows a few failure cases. Majority of the failure cases for digital attacks are due to imperceptible perturbations. In contrast, failure to detect spoofs can likely be attributed to the subtle nature of transparent masks, blurring, and illumination changes.

\Section{Conclusions}
With new and sophisticated attacks being crafted against AFR systems in both digital and physical spaces, detectors need to be robust across all $3$ categories. Prevailing methods indeed excel at detecting attacks in their respective categories, however, they fall short in generalizing across categories. While, ensemble techniques can enhance the overall performance, they still fail to meet the desired accuracy levels. 
Poor generalization can be predominantly attributed towards learning incoherent attacks jointly. 
With a new multi-task learning framework along with $k$-means augmentation, the proposed~\emph{UniFAD} achieved SOTA detection performance (TDR = $94.73\%$ @ $0.2\%$ FDR) on $25$ face attacks across $3$ categories.~\emph{UniFAD} can further identify categories with a $97.37\%$ accuracy. 
We are exploring whether an attention module can further improve detection, or the proposed approach can be applied to generic image manipulation detection beyond faces~\cite{pscc-net-progressive-spatio-channel-correlation-network-for-image-manipulation-detection-and-localization}.

\vspace{2mm}
\noindent\textbf{Acknowledgement}
This work was partially supported by Facebook AI. 

{\small
\bibliographystyle{ieee_fullname}
\bibliography{egbib}
}

\appendix

\section{Details}
Here, we provide additional details on the proposed~\emph{GrandFake} dataset and~\emph{UniFAD} and baselines.

\subsection{GrandFake Dataset} The~\emph{GrandFake} dataset is composed of several widely adopted face datasets for face recognition, face attribute manipulation, and synthesis. Specific details on the source datasets along with training and testing splits are provided in Tab.~\ref{tab:grandfake}. We ensure that there is no identity overlap between any of the training and testing splits as follows:
\begin{enumerate}
    \item We removed $84$ subjects in CASIA-WebFace that overlap with LFW. In addition, CASIA-WebFace is only used for training, while LFW is only used for testing.
    \item SiW-M comprises of high-resolution photos of non-celebrity subjects without any identity overlap with other datasets. Training and testing splits are composed of videos pertaining to different identities.
    \item Identities in CelebA training set are different than those in testing.
    \item FFHQ comprises of high-resolution photos of non-celebrity subjects on Flicker. FFHQ is utilized solely for bona fides in order to add diversity to the quality of face images.
\end{enumerate}

\subsection{Implementation Details}
All the models in the paper are implemented using Tensorflow r1.12. A single NVIDIA GeForce GTX 2080Ti GPU is used for training~\emph{UniFAD} on ~\emph{GrandFake} dataset.
\textbf{Code, pre-trained models and dataset will be publicly available.}

\paragraph{Preprocessing}
All face images are first passed through MTCNN
face detector~\cite{mtcnn} to detect $5$ facial landmarks (two eyes,
nose and two mouth corners). Then, similarity transformation is used to normalize the face images based on the five
landmarks. After transformation, the images are resized to
$160\times 160$. Before passing into~\emph{UniFAD} and baselines, each pixel in the
RGB image is normalized $\in [-1, 1]$ by subtracting $128$ and dividing by $128$. \textbf{All the testing images in the main paper and
this supplementary material are from the identities in the test dataset.} 

\begin{table*}[!t]
\scriptsize
\captionsetup{font=small}
\captionsetup[subfloat]{labelformat=empty}
\centering
\begin{threeparttable}
\renewcommand{\arraystretch}{1.5}
\begin{tabularx}{\linewidth}{l | l | l | M | M | M | M| c}
\noalign{\hrule height 1.0pt}
\multicolumn{2}{l}{} & \textbf{} & \textbf{\# Total Samples} & \textbf{\# Training} & \textbf{\# Validation}  & \textbf{\# Testing} & \textbf{Examples}\\
\noalign{\hrule height 1.0pt}

\parbox[t]{2mm}{\multirow{13}{*}{\raisebox{13.0em}{\rotatebox[origin=c]{90}{Datasets}}}}
&\parbox[t]{2mm}{\multirow{13}{*}{\raisebox{13.0em}{\rotatebox[origin=c]{90}{Bona Fides}}}}

& CASIA-WebFace~\cite{casia} & 10,575 & 10,075 & 500 & 0 &  \multirow{6}{*}{\subfloat[\scriptsize CASIA~\cite{casia}]{\includegraphics[width=0.065\linewidth]{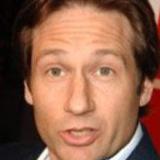}}\quad\subfloat[\scriptsize CelebA~\cite{celeba}]{\includegraphics[width=0.065\linewidth]{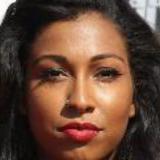}}\quad\subfloat[\scriptsize LFW~\cite{lfw}]{\includegraphics[width=0.065\linewidth]{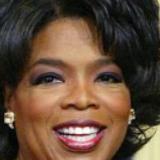}}\quad\subfloat[\scriptsize FFHQ~\cite{ffhq}]{\includegraphics[width=0.065\linewidth]{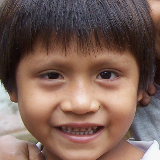}}\quad\subfloat[\scriptsize SiW-M~\cite{siw_m}]{\includegraphics[width=0.065\linewidth]{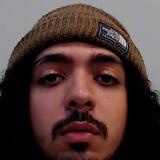}}}\\ \cline{3-7}
&& CelebA~\cite{celeba} & 193,506 & 134,954 & 500 & 58,052  &\\
\cline{3-7}
&& LFW~\cite{lfw} & 9,164 & 0 & 0 & 9,164 &\\
\cline{3-7}
&& FFHQ~\cite{ffhq} & 20,999 & 14,200 & 500 & 6,299 &\\
\cline{3-7}
&& SiW-M~\cite{siw_m} & 107,494 &  74,746 & 500 & 32,248 &\\
\cline{3-7}
&& \textbf{Total} & \mathbf{341,738} & \textbf{233,975} & \textbf{2,000} & \textbf{105,763} &\\
\cline{2-7}

\noalign{\hrule height 1.0pt}
\parbox[t]{2mm}{\multirow{27}{*}{\raisebox{13.0em}{\rotatebox[origin=c]{90}{Attacks}}}}
& \parbox[t]{2mm}{\multirow{13}{*}{\raisebox{13.0em}{\rotatebox[origin=c]{90}{Adversarial}}}}

& FGSM & 19,739 & 10,495  & 80 & 9,164 & \multirow{6}{*}{\subfloat[\scriptsize FGSM]{\includegraphics[width=0.055\linewidth]{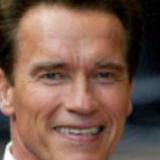}}\quad\subfloat[\scriptsize PGD]{\includegraphics[width=0.055\linewidth]{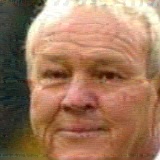}}\quad\subfloat[\scriptsize DeepFool]{\includegraphics[width=0.055\linewidth]{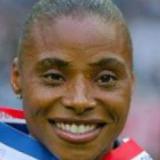}}\quad\subfloat[\scriptsize AdvFaces]{\includegraphics[width=0.055\linewidth]{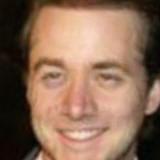}}\quad\subfloat[\scriptsize GFLM]{\includegraphics[width=0.055\linewidth]{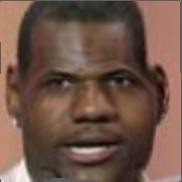}}\quad\subfloat[\scriptsize Semantic]{\includegraphics[width=0.055\linewidth]{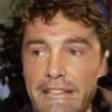}}}\\\cline{3-7}
&& PGD & 19,739 & 10,495 & 80 & 9,164\\\cline{3-7}
&& DeepFool & 19,739 & 10,495 & 80 & 9,164\\\cline{3-7}
&& AdvFaces & 19,739 & 10,495 & 80 & 9,164\\\cline{3-7}
&& GFLM &  17,946 & 8,702 & 80 & 9,164\\\cline{3-7}
&& Semantic & 19,739 & 10,495 & 80 & 9,164\\
\cmidrule[1pt]{2-8}

&\parbox[t]{2mm}{\multirow{11}{*}{\raisebox{13.0em}{\rotatebox[origin=c]{90}{Digital Manipulation}}}}
& DeepFake & 18,165 & 10,393 & 80 & 7,692 & \multirow{6}{*}{\subfloat[\scriptsize DeepFake]{\includegraphics[width=0.055\linewidth]{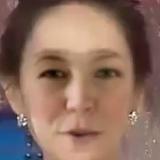}}\quad\subfloat[\scriptsize Face2Face]{\includegraphics[width=0.055\linewidth]{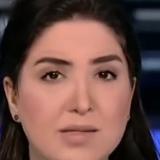}}\quad\subfloat[\scriptsize FaceSwap]{\includegraphics[width=0.055\linewidth]{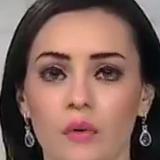}}\quad\subfloat[\scriptsize StarGAN]{\includegraphics[width=0.055\linewidth]{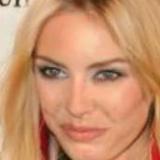}}\quad\subfloat[\scriptsize STGAN]{\includegraphics[width=0.055\linewidth]{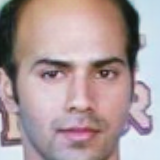}}\quad\subfloat[\scriptsize StyleGAN]{\includegraphics[width=0.055\linewidth]{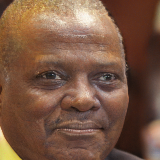}}}\\\cline{3-7}
&& Face2Face & 18,204 & 10,385 & 80 & 7,739\\\cline{3-7}
&& FaceSwap & 14,492 & 8,299 & 80 & 6,113\\\cline{3-7}
&& STGAN & 29,983 & 9,903 & 80 & 20,000\\\cline{3-7}
&& StarGAN & 45,473 & 10,406 & 80 & 34,987\\\cline{3-7}
&& StyleGAN & 76,604 & 10,919 & 80 & 65,605\\
\cmidrule[1pt]{2-8}

&\parbox[t]{2mm}{\multirow{20}{*}{\raisebox{13.0em}{\rotatebox[origin=c]{90}{Spoofs}}}}
& Cosmetic & 2,638 & 1,766 & 80 & 792 & \multirow{4}{*}{\subfloat[\scriptsize Cosm.]{\includegraphics[width=0.055\linewidth]{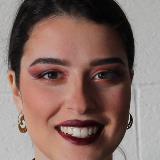}}\quad\subfloat[\scriptsize Imp.]{\includegraphics[width=0.055\linewidth]{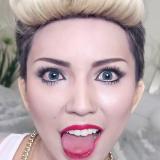}}\quad\subfloat[\scriptsize Obf.]{\includegraphics[width=0.055\linewidth]{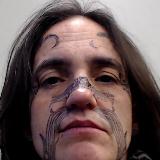}}\quad\subfloat[\scriptsize HalfMask]{\includegraphics[width=0.055\linewidth]{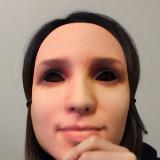}}\quad\subfloat[\scriptsize Mann.]{\includegraphics[width=0.055\linewidth]{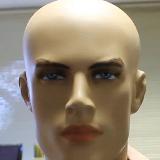}}}\\\cline{3-7}
&& Impersonation & 9,184 & 6,348 & 80 & 2,756\\\cline{3-7}
&& Obfuscation & 3,611 & 2,447 & 80 & 1,084\\\cline{3-7}
&& HalfMask & 10,486 & 7,260 & 80 & 3,146\\\cline{3-7}
&& Mannequin & 5,287 & 3,620 & 80 & 1,587& \multirow{4}{*}{\subfloat[\scriptsize PaperMask]{\includegraphics[width=0.055\linewidth]{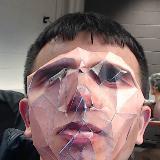}}\quad\subfloat[\scriptsize Silicone]{\includegraphics[width=0.055\linewidth]{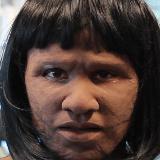}}\quad\subfloat[\scriptsize Trans.]{\includegraphics[width=0.055\linewidth]{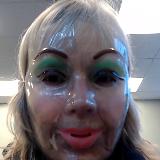}}\quad\subfloat[\scriptsize Print]{\includegraphics[width=0.055\linewidth]{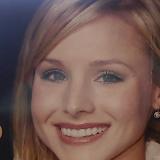}}}\\\cline{3-7}
&& PaperMask & 2,550 & 1,705 & 80 & 765\\\cline{3-7}
&& Silicone & 5,038 & 3,446 & 80 & 1,512\\\cline{3-7}
&& Transparent & 11,451 & 7,935 & 80 & 3,436\\\cline{3-7}
&& Print & 10,530 & 7,290 & 80 & 3,160& \multirow{4}{*}{\subfloat[\scriptsize PaperCut]{\includegraphics[width=0.055\linewidth]{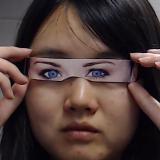}}\quad\subfloat[\scriptsize Funnyeye]{\includegraphics[width=0.055\linewidth]{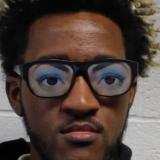}}\quad\subfloat[\scriptsize PaperGlass]{\includegraphics[width=0.055\linewidth]{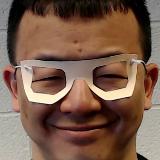}}\quad\subfloat[\scriptsize Replay]{\includegraphics[width=0.055\linewidth]{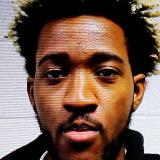}}}\\\cline{3-7}
&& PaperCut & 13,178 & 9,144 & 80 & 3,954\\\cline{3-7}
&& FunnyEye & 23,470 & 16,349 & 80 & 7,041\\\cline{3-7}
&& PaperGlass & 18,563 & 12,914 & 80 & 5,569\\\cline{3-7}
&& Replay & 12,126 & 8,408 & 80 & 3,638\\
\noalign{\hrule height 0.5pt}
\multicolumn{2}{l}{} & \textbf{Total} & \mathbf{447,674} & \textbf{210,114} & \textbf{2,000} & \textbf{235,560} &\\

\noalign{\hrule height 1pt}
\end{tabularx}
 \end{threeparttable}\vspace{-2mm}
\caption{Composition and statistics for the proposed~\emph{GrandFake} dataset. We also include the evaluation protocol for the seen attack scenario (main paper). While SiW-M~\cite{siw_m} is temporarily unavailable, we received a SiW-M-v2 dataset from the authors~\cite{siw_m}, which covers the same 13 different spoof types and will be released to the public as a replacement of SiW-M.}
\label{tab:grandfake}
\end{table*}

\paragraph{Network Architecture}
The backbone network, JointCNN, comprises of a $4$-layer binary CNN:
\begin{itemize}
    \itemsep0em
    \item\texttt{d32,d64,d128,d256,fc128,fc1},
\end{itemize}
where \texttt{dk} denotes a $4\times 4$ convolutional layer with $k$ filters and stride $2$ and $\texttt{fcN}$ refers to a fully-connected layer with $N$ neuron outputs.

The proposed~\emph{UniFAD} with $4$ branches is composed of:
\begin{itemize}
    \itemsep0em
    \item Early Layers: \\ 
    \texttt{d32,d64},
    \item Branch $1$: \\ 
    \texttt{d128,d256,fc128,fc1},
    \item Branch $2$: \\ 
    \texttt{d128,d256,fc128,fc1},
    \item Branch $3$: \\ 
    \texttt{d128,d256,fc128,fc1},
    \item Branch $4$: \\ 
    \texttt{d128,d256,fc128,fc1},
\end{itemize}

\subsection{Digital Attack Implementation}
Adversarial attacks are synthesized via publicly available author codes:
\begin{itemize}[noitemsep]
    \item FGSM/PGD/DeepFool: \url{https://github.com/tensorflow/cleverhans}
    \item AdvFaces: \url{https://github.com/ronny3050/AdvFaces}
    \item GFLM: \url{https://github.com/alldbi/FLM}
    \item SemanticAdv: \url{https://github.com/AI-secure/SemanticAdv}
\end{itemize}
Digital manipulation attacks are also generated via publicly available author codes:
\begin{itemize}[noitemsep]
    \item DeepFake/Face2Face/FaceSwap: \url{https://github.com/ondyari/FaceForensics/tree/original}
    \item STGAN: \url{https://github.com/csmliu/STGAN}
    \item StarGAN: \url{https://github.com/yunjey/stargan}
    \item StyleGAN-v2: \url{https://github.com/NVlabs/stylegan2}
\end{itemize}

\begin{table*}[!t]
\centering
\footnotesize
\captionsetup{font=small}
\begin{threeparttable}
\begin{tabular}{l|lMclMMMMM}
\toprule
\multicolumn{2}{l}{\textbf{Method}} & \textbf{Year} & \textbf{Proposed For} & \textbf{Metric} & \textbf{Adv.} & \textbf{Dig. Man.} & \textbf{Phys.} & \textbf{Overall}\\
\midrule
\parbox[t]{0.3mm}{\multirow{17}{*}{\raisebox{13.0em}{\rotatebox[origin=c]{90}{w/o Re-train}}}}
& \multirow{2}{*}{FaceGuard~\cite{faceguard}} & \multirow{2}{*}{2020} & \multirow{2}{*}{Adversarial} & TDR & \firstkey{99.91} & 22.28 & 00.58 & 29.64\\
& & & & Acc. & \firstkey{99.78} & 67.12 & 51.02 & 71.03\\ \cline{2-9}
& \multirow{2}{*}{FFD~\cite{detection_digital}} & \multirow{2}{*}{2020} & \multirow{2}{*}{Digital Manipulation} & TDR & 09.49 & \secondkey{94.57} & 01.25 & 34.55\\
& & & & Acc. & 56.42 & \secondkey{97.87} & 53.42 & 75.29\\\cline{2-9}
& \multirow{2}{*}{SSRFCN~\cite{ssrfcn}} & \multirow{2}{*}{2020} & \multirow{2}{*}{Spoofs} & TDR & 00.25 & 00.76 & 93.19 & 22.71\\
& & & & Acc. & 50.01 & 50.93 & 96.12 & 69.11\\\cline{2-9}
& \multirow{2}{*}{MixNet~\cite{mixnet}} & \multirow{2}{*}{2020} & \multirow{2}{*}{Spoofs} & TDR & 00.36 & 09.83 & 78.21 & 21.12\\
&  & &  & Acc. & 50.43 & 55.98 & 85.47 & 61.26\\
\midrule
\parbox[t]{0.3mm}{\multirow{20}{*}{\raisebox{13.0em}{\rotatebox[origin=c]{90}{Baselines}}}}
& \multirow{2}{*}{FaceGuard~\cite{faceguard}} & \multirow{2}{*}{2020} & \multirow{2}{*}{Adversarial}  & TDR & \secondkey{99.86} & 41.56 & 04.35 & 56.69\\
& & & & Acc. & \secondkey{99.71} & 71.23 & 54.06 & 81.88\\\cline{2-9}
& \multirow{2}{*}{FFD~\cite{detection_digital}}  & \multirow{2}{*}{2020} & \multirow{2}{*}{Digital Manipulation}  & TDR & 76.06 & 91.32 & 87.43 & 68.25\\
& & & & Acc. & 87.15 & 93.40 & 91.37 & 89.06\\\cline{2-9}
& \multirow{2}{*}{SSRFCN~\cite{ssrfcn}} & \multirow{2}{*}{2020} & \multirow{2}{*}{Spoofs}  & TDR & 08.23 & 27.67 & 89.19 & 43.26 \\
& & & & Acc. & 54.02 & 69.18 & 90.91 & 83.41\\\cline{2-9}
& \multirow{2}{*}{One-class~\cite{one_class_spoof}} & \multirow{2}{*}{2020} & \multirow{2}{*}{Spoofs}  & TDR & 04.81 & 45.96 & 79.32 & 39.40\\
&  &  &  & Acc. & 53.99 & 64.08 & 86.65 & 80.74\\\cline{2-9}
& \multirow{2}{*}{MixNet-\emph{UniFAD}} & \multirow{2}{*}{2021} & \multirow{2}{*}{All} & TDR & 82.33 & 91.59 & \secondkey{94.60} & \secondkey{90.07}\\
& &  & & Acc. & 89.32 & 94.50 & \secondkey{96.18} & \secondkey{93.19}\\
\midrule
\parbox[t]{0.3mm}{\multirow{23}{*}{\raisebox{13.0em}{\rotatebox[origin=c]{90}{Fusion Schemes}}}}
& \multirow{2}{*}{Cascade~\cite{viola_jones}} & \multirow{2}{*}{-} & \multirow{2}{*}{$-$} & TDR & 88.39 & 81.98 & 69.19 & 77.46\\
& & &  & Acc. & 91.33 & 89.17 & 79.92 & 85.16\\\cline{2-9}
& \multirow{2}{*}{Min-score} & \multirow{2}{*}{-} & \multirow{2}{*}{$-$} & TDR & 03.65 & 11.08 & 00.43 & 07.22\\
& & & & Acc. & 51.61 & 66.76 & 50.88 & 55.62\\\cline{2-9}
& \multirow{2}{*}{Median-score} & \multirow{2}{*}{-} & \multirow{2}{*}{$-$} & TDR & 10.87 & 42.33 & 47.19 & 39.48\\
& & & & Acc. & 55.12 & 59.58 & 57.44 & 59.22\\ \cline{2-9}
& \multirow{2}{*}{Mean-score} & \multirow{2}{*}{-} & \multirow{2}{*}{$-$} & TDR & 14.53 & 47.18 & 61.32 & 38.23\\
&  & & & Acc. & 55.69 & 54.19 & 73.87 & 55.92\\\cline{2-9}
& \multirow{2}{*}{Max-score} & \multirow{2}{*}{-} & \multirow{2}{*}{$-$} & TDR & 85.32 & 61.93 & 56.87 & 73.89\\
& & & & Acc. & 89.26 & 68.11 & 60.08 & 69.43\\\cline{2-9}
& \multirow{2}{*}{Sum-score} & \multirow{2}{*}{-} & \multirow{2}{*}{$-$} & TDR & 74.93 & 58.01 & 50.34 & 69.21\\
& & & & Acc. & 83.85 & 67.48 & 64.72 & 73.10\\\cline{2-9}
& \multirow{2}{*}{LightGBM~\cite{lightgbm}} & \multirow{2}{*}{-} & \multirow{2}{*}{$-$} & TDR & 76.25 & 81.28 & 88.52  & 85.97\\
& & & & Acc. & 84.19 & 89.46 & 94.56  & 90.56\\
\midrule
\multicolumn{1}{l}{} & \multirow{2}{*}{\emph{Proposed UniFAD}} & \multirow{2}{*}{2021} & \multirow{2}{*}{All} & TDR & 92.56 & \firstkey{97.21} & \firstkey{98.76} & \firstkey{94.73}\\
\multicolumn{1}{l}{} & & & & Acc & 95.18 & \firstkey{98.32} & \firstkey{98.96} & \firstkey{96.89}\\
\bottomrule
\end{tabular}
\end{threeparttable}\vspace{-2mm}
\caption{Detection performance (TDR (\%) @ $0.2\%$ FDR and Accuracy (\%)) on~\emph{GrandFake} dataset under the \emph{seen} attack scenario. [Keys: \firstkeytext{Best}, \secondkeytext{Second best}]}
\label{tab:detection_seen}
\end{table*}

\subsection{Baseline Implementation}
\ParagraphMini{Individual Detectors} We evaluate all~\emph{individual} defense methods, except MixNet~\cite{mixnet}, via publicly available repositories provided by the authors. We provide the public links to the author codes below:
\begin{itemize}[noitemsep]
    \item FaceGuard~\cite{faceguard}: \url{https://github.com/ronny3050/FaceGuard}
    \item FFD~\cite{detection_digital}: \url{https://github.com/JStehouwer/FFD_CVPR2020}
    \item SSRFCN~\cite{ssrfcn}: \url{https://github.com/ronny3050/SSRFCN}
    \item One-Class~\cite{one_class_spoof}: \url{https://github.com/anjith2006/bob.paper.oneclass_mccnn_2019}
\end{itemize}

\ParagraphMini{Fusion of JointCNNs} In our work, we employ $5$ parallel score-level fusion rules. For a testing input image, we extract three scores (in $[0, 1]$) from three SOTA individual detectors (FaceGuard~\cite{faceguard}, FFD~\cite{detection_digital}, and SSRFCN~\cite{ssrfcn}). The final decision score is computed via the fusion rule operator, namely~\emph{min},~\emph{mean},~\emph{median},~\emph{max}, and~\emph{sum}. LightGBM~\cite{lightgbm} is a tree-ensemble learning method where a Gradient Boosted Decision Tree is trained from the three scores (from individual SOTA dectors) to output to the final decision. We use Microsoft's LightGBM implementation:~\url{https://github.com/microsoft/LightGBM}.~\textbf{We train the LightGBM model on the training set of~\emph{GrandFake}.}

\section{Additional Experiments} In this section, we compare the proposed~\emph{UniFAD} to prevailing face attack detectors under the~\emph{seen} attack scenario and also generalization performance under unseen attacks types.

\subsection{Seen Attacks} 
Tab.~\ref{tab:detection_seen} reports the detection performance (TDR (\%) @ $0.2\%$ FDR and accuray (\%)) of~\emph{UniFAD} and baselines on~\emph{GrandFake} dataset under the seen attack scenario. Training and testing splits are provided in Tab.~\ref{tab:grandfake}. Overall,~\emph{UniFAD} outperforms all fusion schemes and baselines. 

\begin{figure*}
    \centering
    \captionsetup{font=small}
    \includegraphics[width=0.9\linewidth]{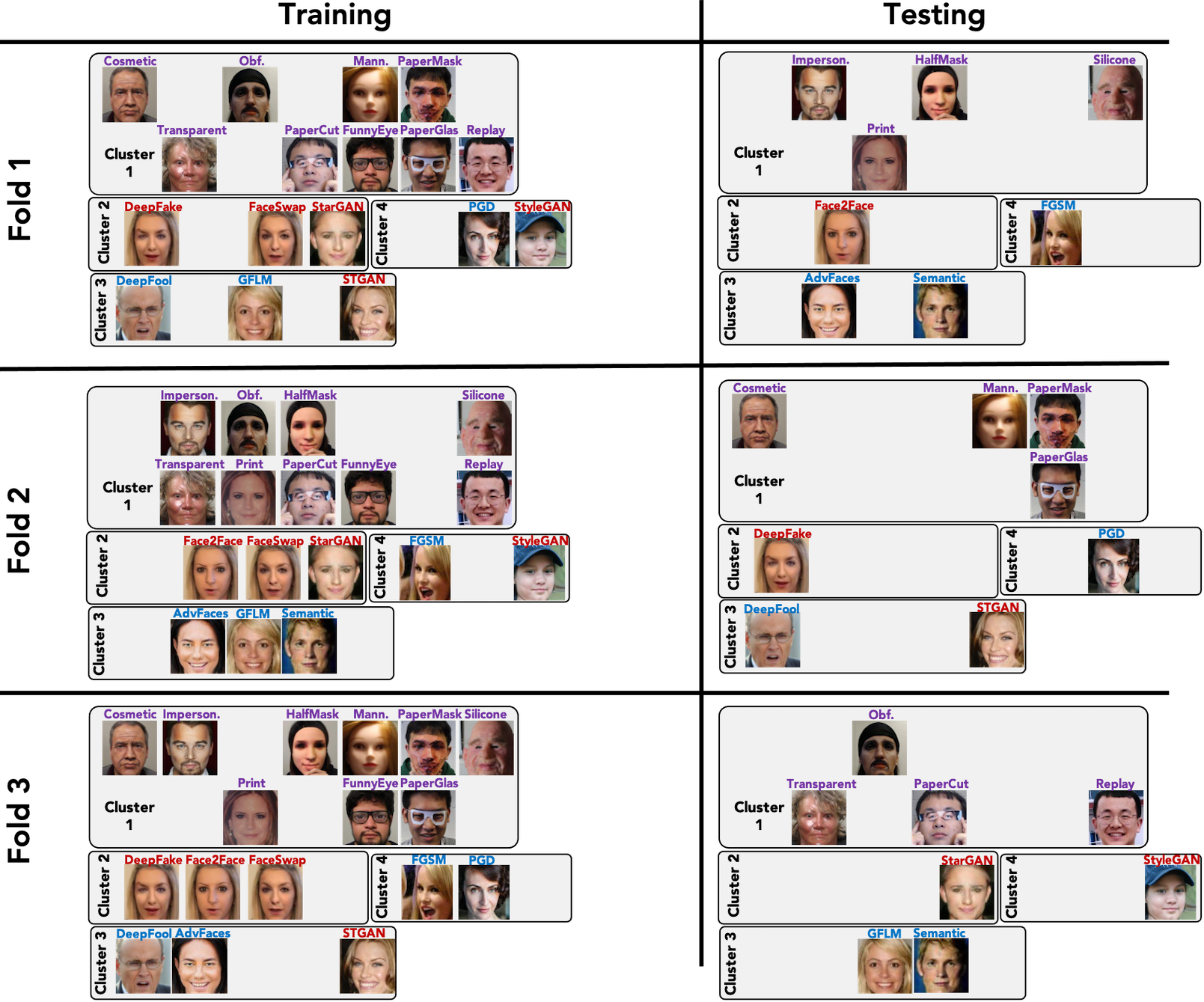}
    \caption{Training and testing splits for generalizability study.}
    \label{fig:protocols}
\end{figure*}

\begin{table}[!h] 
\captionsetup{font=small}
\centering
\def\arraystretch{1.5}
\resizebox{\textwidth}{!}{\begin{tabular}{c||c|||M||M||M||M}
\noalign{\hrule height 1.5pt}
\textbf{Method} & \textbf{Metric (\%)} & \textbf{Fold 1} & \textbf{Fold 2} & \textbf{Fold 3} &\textbf{Mean $\pm$ Std.}\\ \hline
\hline\hline
\multirow{2}{*}{FaceGuard~\cite{faceguard}} 
& TDR & 41.38 & 54.19 & 36.82 & 44.13 \pm 9.01\\
& Acc. & 58.42 & 64.19 & 55.74  &  59.45\pm 4.32 \\
\noalign{\hrule height 0.8pt}
\multirow{2}{*}{FFD~\cite{detection_digital}} 
& TDR &  53.19 & 62.45 & 52.94  & 56.20 \pm 5.42\\
& Acc. & 66.15 & 69.33 & 67.86  & 67.78\pm 1.59\\
\noalign{\hrule height 0.8pt}
\multirow{2}{*}{SSRFCN~\cite{ssrfcn}} 
& TDR &  49.10 & 64.92  & 61.18  & 58.84 \pm 8.26 \\
& Acc. &  60.07 & 72.77  & 69.83 & 66.57 \pm 6.64\\
\noalign{\hrule height 0.8pt}
\multirow{2}{*}{MixNet-~\emph{UniFAD}} 
& TDR & \secondkey{67.19} & \secondkey{73.18} & \secondkey{72.74}  & \secondkey{71.04 \pm 3.33}\\
& Acc. & \secondkey{75.64}  & \secondkey{79.40}  & \secondkey{78.73}  & \secondkey{77.93 \pm 2.00}\\
\noalign{\hrule height 0.8pt}
\multirow{2}{*}{LightGBM} 
& TDR &  51.65 & 65.73 & 67.91 & 61.76 \pm 8.83\\
& Acc. &  69.34 & 73.66 & 75.80 & 72.93 \pm 3.29 \\
\noalign{\hrule height 0.8pt}
\multirow{2}{*}{Proposed~\emph{UniFAD}} 
& TDR & \firstkey{76.18} & \firstkey{83.19} & \firstkey{82.67} & \firstkey{80.68 \pm 3.91}\\
& Acc. &  \firstkey{85.35} & \firstkey{89.62} & \firstkey{85.88} & \firstkey{86.95 \pm 2.32}\\
\noalign{\hrule height 1.5pt}
\end{tabular}}
\vspace{-2mm}\caption{Generalization performance (TDR (\%) @ $0.2\%$ FDR and Accuracy (\%)) on~\emph{GrandFake} dataset under unseen attack setting.  Each fold comprises of $8$ unseen attacks from all 4 branches.
[Keys: \firstkeytext{Best}, \secondkeytext{Second best}]}
\label{tab:unseen}
\end{table}



\subsection{Generalizability to Unseen Attacks} Under this setting, we evaluate the generalization performance on 3 folds (see Fig.~\ref{fig:protocols}). The folds are computed as follows: we hold out $1/3$ of the total attack types in a branch for testing and the remaining are used for training. For~\eg, branch 1 consisting of $13$ attack types are~\emph{randomly} split such that we test on $4$ unseen attack types, while the remaining $9$ attack types are used for training. We perform $3$ folds of such random splitting. In total, each fold consists of $17$ seen and $8$ unseen attacks.  For LightGBM, we utilize scores from FaceGuard~\cite{faceguard}, FFD~\cite{detection_digital}, and SSRFCN~\cite{ssrfcn} which are all trained only on the known attack types. We report the detection performance and the average and standard deviation across all folds in Tab.~\ref{tab:unseen}. 

\begin{figure}
    \captionsetup{font=small}
    \centering
    \includegraphics[width=0.85\linewidth]{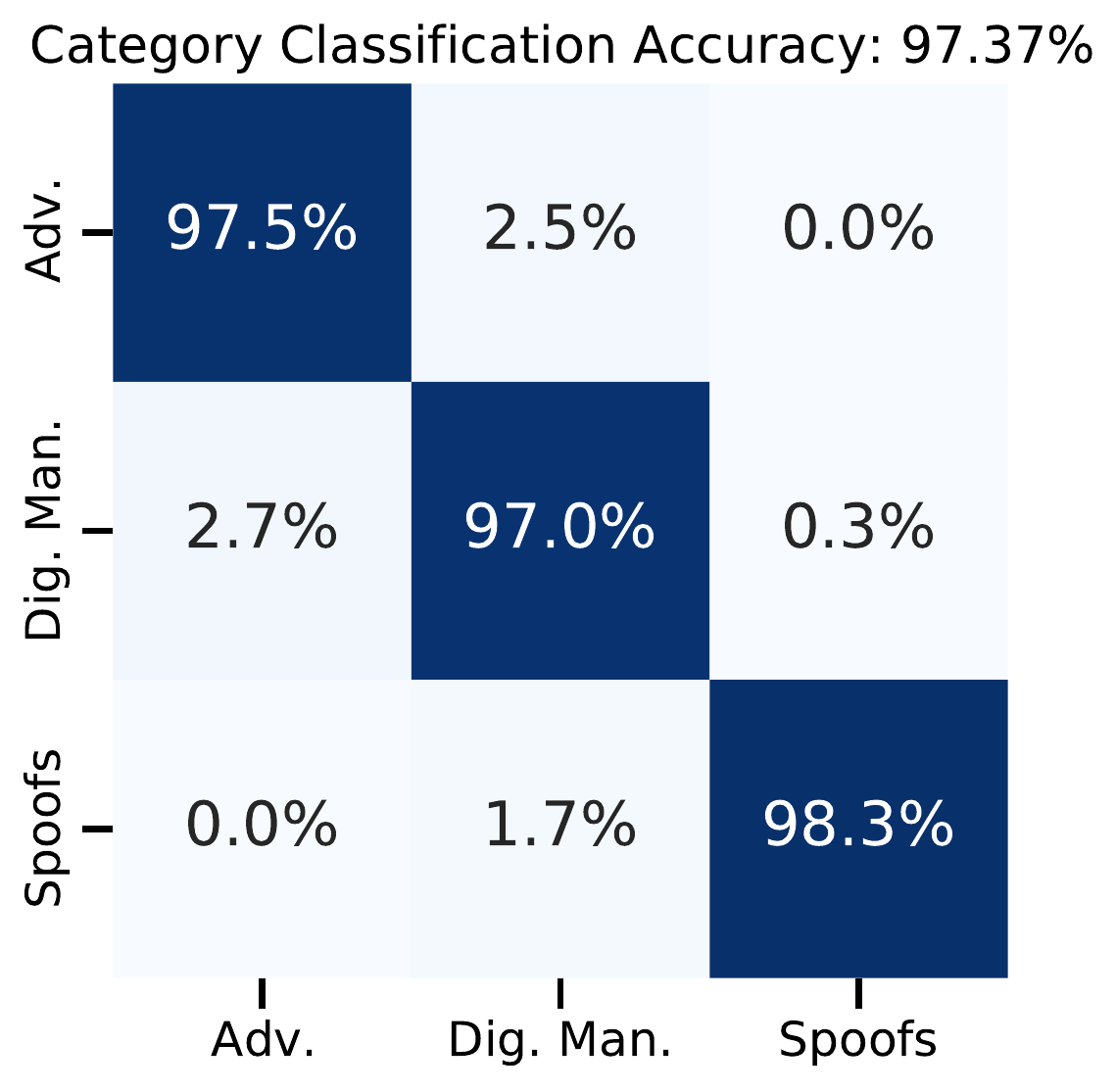}\label{fig:cm_cat}\vspace{-2mm}
    \caption{Confusion matrix representing the classification accuracy of~\emph{UniFAD} in identifying the $3$ attack categories, namely adversarial faces, digital face manipulation, and spoofs. Majority of confusion occurs within digital attacks (adversarial and digital manipulation attacks).}
    \label{fig:classification_cat}
\end{figure}

We find that branching-based methods, such as MixNet-~\emph{UniFAD} and the proposed~\emph{UniFAD}, significantly outperforms JointCNN-based methods such as FaceGuard~\cite{faceguard}, FFD~\cite{detection_digital}, SSRFCN~\cite{ssrfcn}, and LightGBM (fusion of the three). The superiority of the proposed~\emph{UniFAD} under unseen attacks is evident. By incorporating branches with coherent attacks, removing some attack types within a branch does not drastically affect the generalization performance. 

In addition to superior generalization performance to unseen attack types, the proposed~\emph{UniFAD} also reduces the gap between seen and unseen attacks. Overall, the proposed~\emph{UniFAD} achieves $94.73\%$ and $80.68\%$ TDRs at $0.2\%$ FDR under seen and unseen attack scenario. That is, we observe a relative reduction in TDR of $15\%$ under unseen attacks, compared to the second best method, namely MixNet-\emph{UniFAD}, which has a relative reduction in TDR of $22\%$ under unseen attacks.

\section{Attack Category Classification} 
In Fig.~\ref{fig:classification_cat}, we find that~\emph{UniFAD} can predict the attack category with $97.37\%$ classification accuracy. We emphasize that majority of the misclassifications occurs within the digital attack space. That is, misclassifying adversarial attacks as digital manipulation attacks and vice-versa. 

Among spoofs, $1.7\%$ of them are misclassified as digital manipulation attacks. Majority of these are makeup attacks which have some correlation with some digital manipulation attacks such as DeepFake, Face2Face, and FaceSwap (see Fig. \textcolor{red}{2} in \emph{main paper}). We posit that this likely because cosmetic and impersonation attacks apply makeup to eyebrows and cheeks which may appear similar to ID-swapping methods such as DeepFake and Face2Face which also majorly alter the eyebrows and cheeks.

\end{document}